\documentclass[10pt,journal,compsoc]{utils/IEEEtran}

\ifCLASSOPTIONcompsoc
  \usepackage[nocompress]{cite}
\else
  \usepackage{cite}
\fi

\usepackage{amssymb}
\usepackage{graphicx}
\usepackage{subfigure}
\usepackage{amsmath}
\usepackage{url}
\usepackage{color}
\usepackage{bbm}
\usepackage{makecell, verbatim, sidecap}
\usepackage{balance}
\usepackage[pagebackref,breaklinks,colorlinks]{hyperref}

\def\eg{{\em e.g.}}
\def\ie{{\em i.e.}}

\begin{document}
\title{CenterNet++ for Object Detection}
%
%
%
%

\author{Kaiwen~Duan, Song~Bai, Lingxi~Xie, Honggang~Qi, Qingming~Huang
and~Qi~Tian
\IEEEcompsocitemizethanks{\IEEEcompsocthanksitem Kaiwen Duan, Honggang Qi and Qingming Huang are with the School of Computer Science
and Technology, University of Chinese Academy of Sciences, Beijing, China.
\IEEEcompsocthanksitem Song~Bai is with Huazhong University of Science and Technology.
\IEEEcompsocthanksitem Lingxi~Xie and Qi~Tian are with Huawei Inc..
\IEEEcompsocthanksitem Honggang~Qi and Qingming~Huang are the
corresponding authors (e-mail: \{hgqi,qmhuang\}@ucas.ac.cn).}}
\IEEEtitleabstractindextext{%
\begin{abstract}
There are two mainstreams for object detection: top-down and bottom-up. The state-of-the-art approaches mostly belong to the first category. In this paper, we demonstrate that the bottom-up approaches are as competitive as the top-down and enjoy higher recall. Our approach, named CenterNet, detects each object as a triplet keypoints (top-left and bottom-right corners and the center keypoint). We firstly group the corners by some designed cues and further confirm the objects by the center keypoints. The corner keypoints equip the approach with the ability to detect objects of various scales and shapes and the center keypoint avoids the confusion brought by a large number of false-positive proposals. Our approach is a kind of anchor-free detector because it does not need to define explicit anchor boxes. We adapt our approach to the backbones with different structures, {\em i.e.}, the `hourglass' like networks and the the `pyramid' like networks, which detect objects on a single-resolution feature map and multi-resolution feature maps, respectively. On the MS-COCO dataset, CenterNet with Res2Net-101 and Swin-Transformer achieves APs of $53.7\%$ and $57.1\%$, respectively, outperforming all existing bottom-up detectors and achieving state-of-the-art. We also design a real-time CenterNet, which achieves a good trade-off between accuracy and speed with an AP of $43.6\%$ at $30.5$ FPS. \url{https://github.com/Duankaiwen/PyCenterNet}.
\end{abstract}

}

\maketitle

\IEEEdisplaynontitleabstractindextext
\IEEEpeerreviewmaketitle

\begin{figure}[!t]
  \centering
  \subfigure[]{ 
    \includegraphics[height=0.155\textwidth, width = 0.22\textwidth]{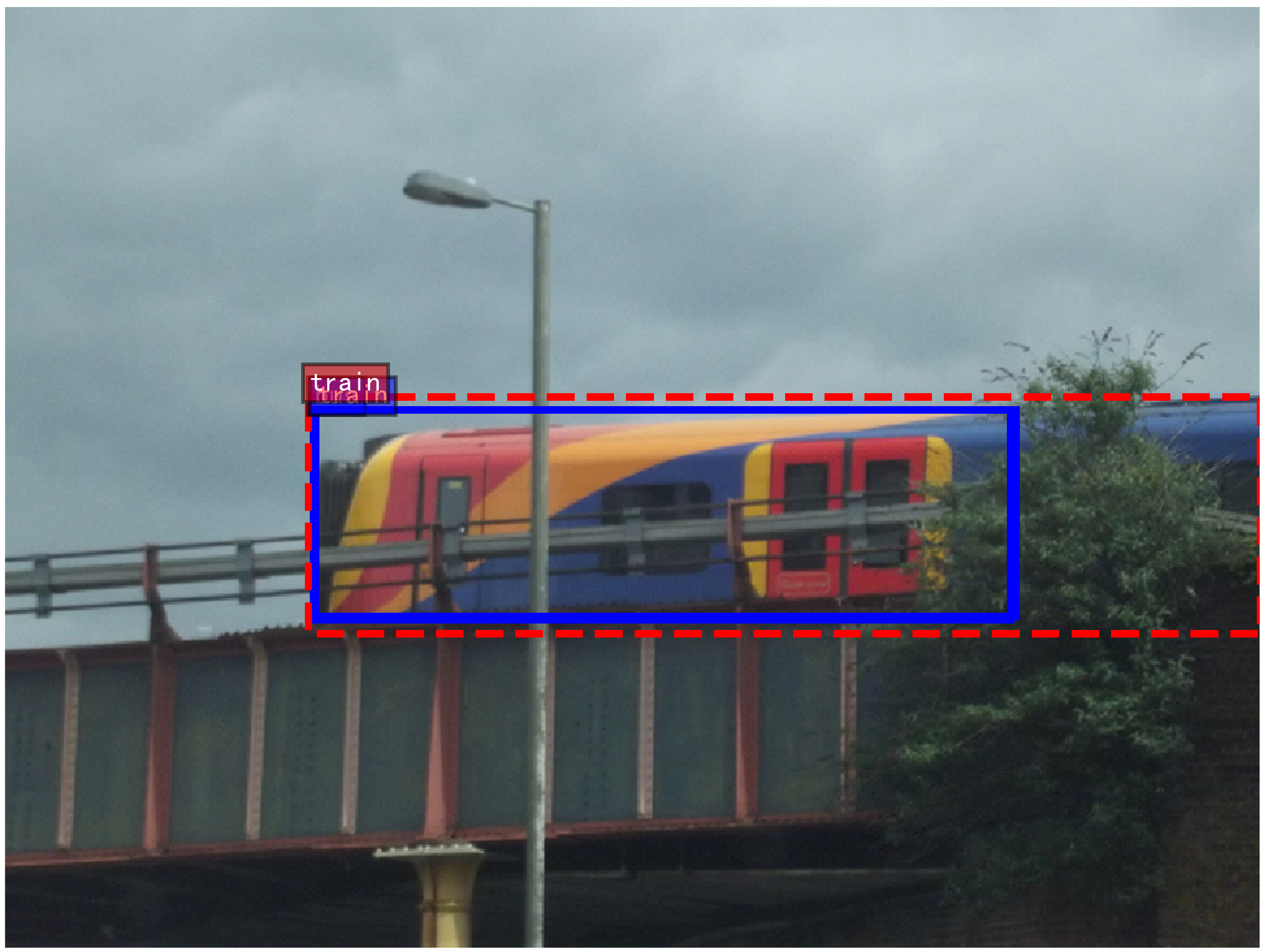}
    \label{fig:faster_result}
  } 
  \hspace{1ex}
  \subfigure[]{ 
    \includegraphics[height=0.155\textwidth, width = 0.215\textwidth]{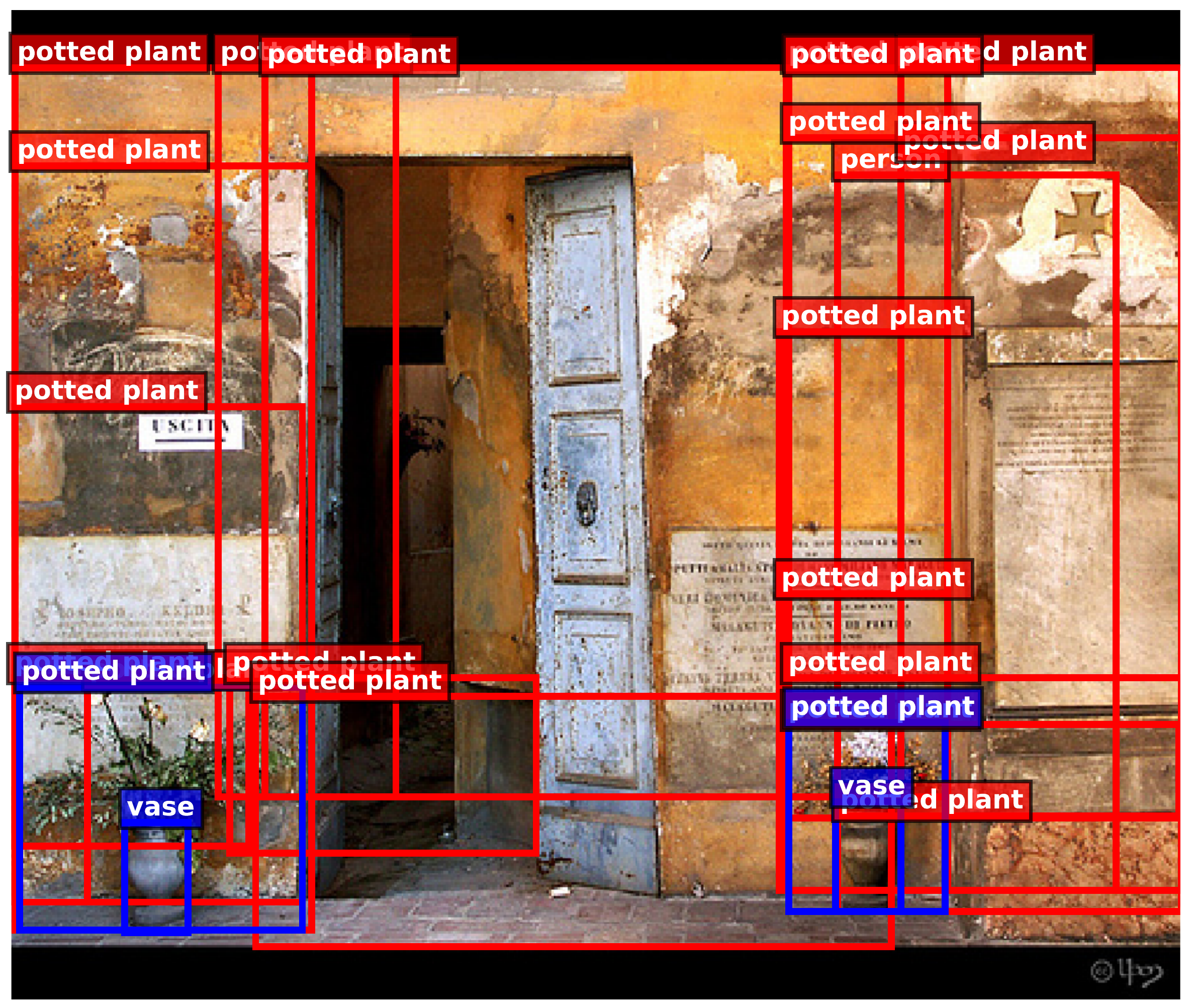}
    \label{fig:cornernet_result} 
  } 
  \subfigure[]{ 
    \includegraphics[height=0.16\textwidth, width = 0.46\textwidth]{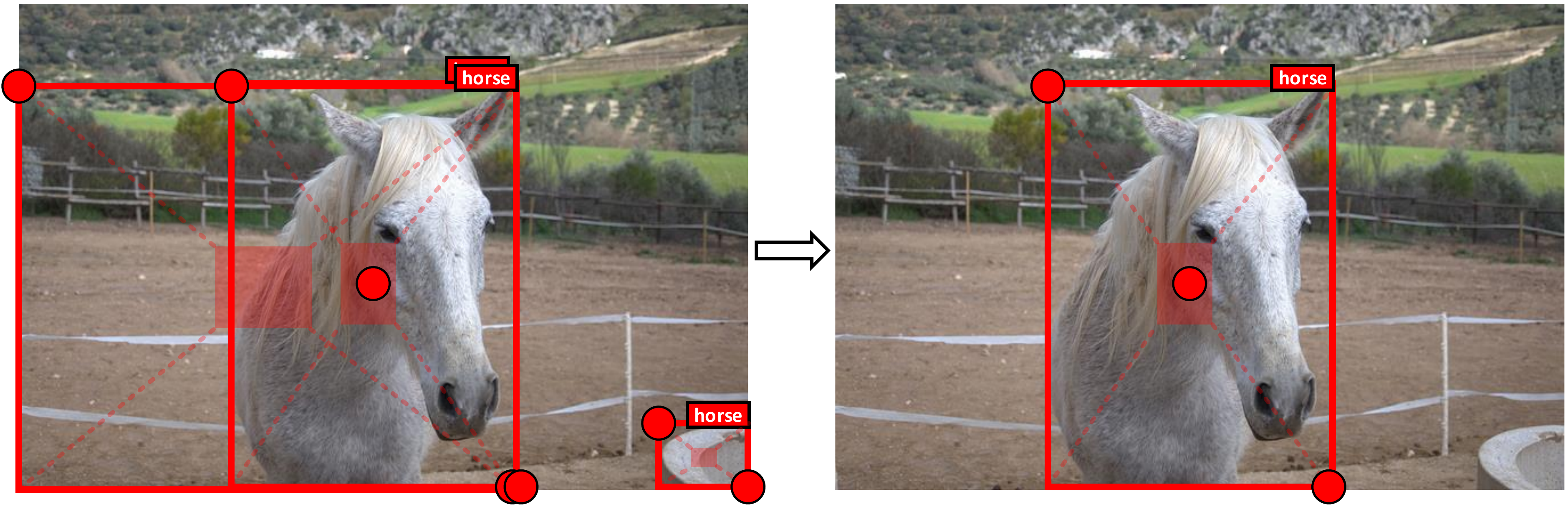}
    \label{fig:centernet_principle} 
  }
  \caption{(a) The top-down approach Faster R-CNN~\cite{ren2016faster} often fails to locate the object precisely that has a peculiar shape (an extreme aspect ratio). Blue and red bounding-boxes indicate false positives and false negatives, respectively. (b) We visualize the top 100 bounding boxes (according to the MS-COCO dataset standard) of CornerNet~\cite{law2018cornernet}, which shows a large number of false positives. Blue and red bounding-boxes indicate true positives and false positives, respectively. (c) The idea of the proposed CenterNet. We show that correct predictions can be determined by checking the central parts.}
  \label{fig:introduction}
\end{figure}

\IEEEraisesectionheading{\section{Introduction}\label{sec:introduction}}
In the current era, there mainly exist two categories for object detection: the bottom-up detection approaches~\cite{zhou2019bottom, law2019cornernet, law2018cornernet, dong2020centripetalnet} and the top-down approaches~\cite{liu2016ssd,redmon2017yolo9000,dai2016r,dai2017deformable}. People believed that the bottom-up approaches may be time-consuming and introduces more false positives, while the top-down approaches have gradually evolved into the mainstream approaches due to their effectiveness in practice. All the top-down approaches model each object as a prior point or a pre-defined anchor box, then predicts the corresponding offsets to the bounding box. They enjoy the ability to perceive the objects as a whole, which simplify the post-process of generating bounding boxes. However, they usually suffer the difficulty to perceive objects with peculiar shapes (\eg, the aspect ratios of the objects are large). Fig.~\ref{fig:faster_result} shows a case that a top-down approach fails to cover the 'train'. We will give a detailed analysis for this problem in Section~\ref{subsec:baseline}.

On the other hand, we find that the bottom-up approaches are potentially better in locating objects with arbitrary geometry, and thus have a higher recall. But the traditional bottom-up approaches usually generate many false positives as well, which fails to represent objects accurately. Take the CornerNet~\cite{law2018cornernet} as an example, one of the most representative bottom-up approaches, it models each object using a pair of corner keypoints and achieves state-of-the-art object detection accuracy. Nevertheless, the performance of CornerNet is still restricted by its relatively weak ability to refer to the global information of an object. That is, because each object is constructed by a pair of corners, the algorithm sensitively detects the boundaries of objects without being aware of which pairs of keypoints that should be grouped into objects. Consequently, as shown in Fig.~\ref{fig:cornernet_result}, CornerNet often generates incorrect bounding boxes, most of which could be easily filtered out with some complementary information, \eg, the aspect ratio. 

Driven by the analysis of the bottom-up approaches, our core opinion is that \textbf{the bottom-up approaches would be necessary and be as competitive as the top-down approaches, as long as we improve their ability to perceive the global information of an object}. In this paper, we present a low-cost yet effective solution named CenterNet, a strong bottom-up object detection approach, that detects each object as a triplet keypoints (top-left and bottom-right corners and the center keypoint). CenterNet explores the central part of a proposal, \ie, the region that is close to the geometric center of a box, with one extra keypoint. We intuit that if a predicted bounding box has a high IoU with the ground-truth box, then the probability that the center keypoint in the central region of the bounding box will be predicted as the same class is high, and vice versa. Thus, during inference, after a proposal is generated as a pair of corner keypoints, we determine if the proposal is indeed an object by checking if there is a center keypoint of the same class falling within its central region. The idea is shown in Fig.~\ref{fig:centernet_principle}. 

We design two frameworks to adapt to the networks with different structures, the first could fit the 'hourglass' like networks, which detect objects on a single-resolution feature map. This kind of networks are very popular in the keypoint estimation task, \eg, the hourglass network~\cite{newell2016stacked}, which we apply the network to better predict the corners and center keypoints. We also design our framework to fit the 'pyramid' like networks, which detect objects on multi-resolution feature maps. This brings two advantages: Stronger generality--most of networks have 'pyramid' structures, \eg, the ResNet~\cite{he2016deep} and its variants; Higher detection accuracy--objects with different scales are detected in different receptive fields. Although the pyramid structure has been widely applied in the top-down approaches, it is, as far as we know, the first time that used in the bottom-up approaches.

We evaluate the proposed CenterNet on the MS-COCO dataset~\cite{lin2014microsoft}, one of the most popular benchmarks for large-scale object detection. CenterNet, with Res2Net-101~\cite{gao2019res2net} and Swin-Transformer~\cite{liu2021swin}, achieves APs of $53.7$ and $57.1$, respectively, outperforming all existing bottom-up detectors by a large margin. We also design a real-time CenterNet, which achieves a good trade-off between accuracy and speed with an AP of $43.6$ at $30.5$ FPS. CenterNet is quite efficient yet closely matches state-of-the-art performance of the other top-down approaches.

The preliminary version of this paper appeared as~\cite{duan2019centernet}. In this extended journal version, we improve the work from the following aspects. (i) The original CenterNet is only applied to the Hourglass network~\cite{newell2016stacked} as the backbone, which all the objects are only detected in a single-resolution feature map. We extend the idea of CenterNet to make it work in the network that has the pyramid structure, which allows CenterNet detecting objects in multi-resolution feature maps. To this end, we propose new methods for detecting keypoints (including corners and center keypoints) and grouping keypoints, respectively. (ii) In this version, due to our new design of CenterNet, we could try more backbones that have the pyramid structures, such as ResNet~\cite{he2016deep}, ResNext~\cite{xie2017aggregated} and Res2Net~\cite{gao2019res2net}. We even report the detection results using the backbone of the Transformer~\cite{liu2021swin}. The experimental results show the detection accuracy is significantly improved by introducing the pyramid structure, which allow the network using the richer receptive fields to detect objects. (iii) We present a real-time CenterNet in this journal version, which achieves a better accuracy/speed trade-off among the popular detectors.

The main contributions of this work can be summarized as follows:
\begin{itemize}
\item We propose a strong bottom-up object detection approach named CenterNet. It detects each object as a triplet keypoints, which enjoys the ability in locating objects with arbitrary geometry and to perceive the global information within objects.
\item We design two frameworks to adapt to the networks with different structures, which enjoys a stronger generality: Our approach could fit almost all of networks. 
\item Without bells and whistles, CenterNet achieves state-of-the-art detection accuracy among the bottom-up approaches and closely matches state-of-the-art performance of the other top-down approaches.
\item With properly reduced the structure complexity, CenterNet achieves a satisfying trade-off between accuracy and speed. We demonstrate that the bottom-up approaches are necessary and as competitive as the top-down approaches.
\end{itemize}

The remainder of this paper is organized as follows. Section~\ref{sec:relatedWork} briefly reviews related work, and Section~\ref{sec:approach} details the proposed CenterNet. Experimental results are given in Section~\ref{sec:experiments}, followed by the conclusion in Section~\ref{sec:conclusions}.

\section{Related Work}\label{sec:relatedWork}

Object detection involves locating and classifying the objects. In the deep learning era, powered by deep convolutional neural networks, object detection approaches can be roughly categorized into two main types of pipelines, namely, top-down approaches and bottom-up approaches. 

\noindent \textbf{Top-down approaches}~first find the proposals that represent the whole objects, then further determine the classes and the bounding boxes of the objects by classifying and regressing the proposals. The proposals could be further divided into \textit{anchor-based} and \textit{anchor-free} according to the different forms of proposals. 

\begin{table*}[!t]
\begin{center}
\caption{The average recall (AR) of top-down and bottom-up approaches on the MS coco validation dataset. In this experiment, we report the AR computed by targets of different aspect ratios and different sizes. To eliminate the influence of other factors on AR, we exclude the impacts of bounding-box categories and scores on recall and compute it by allowing at most $1000$ object proposals. $\mathrm{AR_{1+}}$, $\mathrm{AR_{2+}}$, $\mathrm{AR_{3+}}$ and $\mathrm{AR_{4+}}$ denote box area in the ranges of $\left(96^{2}, 200^{2}\right]$, $\left(200^{2}, 300^{2}\right]$, $\left(300^{2}, 400^{2}\right]$, and $\left(400^{2}, +\infty\right)$, respectively. `X' and `HG' denote ResNeXt and Hourglass, respectively.}
\label{tab:false_negatives}
\resizebox{0.9\textwidth}{!}{
\begin{tabular}{|l|c|c|cccc|cccc|}
\hline
Method & Backbone & $\mathrm{AR}$ & $\mathrm{AR_{1+}}$ & $\mathrm{AR_{2+}}$ & $\mathrm{AR_{3+}}$ & $\mathrm{AR_{4+}}$ & $\mathrm{AR_{5:1}}$ & $\mathrm{AR_{6:1}}$ & $\mathrm{AR_{7:1}}$ & $\mathrm{AR_{8:1}}$\\
\hline
\textbf{Top-down:} & & & & & & & & & & \\
Faster R-CNN~\cite{ren2016faster} & X-101-64x4d & 57.6 & 73.8 & 77.5 & 79.2 & 86.2 & 43.8 & 43.0 & 34.3 & 23.2\\
FCOS~\cite{tian2020fcos} & X-101-64x4d & 64.9 & 82.3 & 87.9 & 89.8 & 95.0 & 45.5 & 40.8 & 34.1 & 23.4\\
\hline
\textbf{Bottom-up} & & & & & & & & & & \\
CornerNet~\cite{law2018cornernet} & HG-104 & 66.8 & 85.8 & 92.6 & 95.5 & 98.5 & 50.1 & 48.3 & 40.4 & 36.5\\
CenterNet (this work) & HG-104 & 66.8 & 87.1 & 93.2 & 95.2 & 96.9 & 50.7 & 45.6 & 40.1 & 32.3\\
\hline
\end{tabular}}
\end{center}
\end{table*}

\textit{Anchor-based} proposals, which are also called anchors, start with rectangles that have different predefined sizes, scales and shapes. They are uniformly distributed on the feature maps and are trained to regress to the desired place with the help of ground-truth objects. Among which, some approaches pay more attention to the quality of detection results. One of the most representative approaches is R-CNN~\cite{girshick2014rich}. It divides the process of object determination into two stages. Some meaningful proposals are selected out in the first stage, and are further verified in the second stage. After that, a lot of works are proposed to expand it, such as SPPNet~\cite{he2015spatial}, Fast R-CNN~\cite{girshick2015fast}, Faster R-CNN~\cite{ren2016faster}, Cascade R-CNN~\cite{cai2018cascade}, MR-CNN~\cite{gidaris2015object}, ION~\cite{bell2016inside}, OHEM~\cite{shrivastava2016training}, HyperNet~\cite{kong2016hypernet}, CRAFT~\cite{yang2016craft}, R-FCN~\cite{dai2016r}, FPN~\cite{lin2017feature},  Libra R-CNN~\cite{pang2019libra}, Mask R-CNN~\cite{he2018mask}, Fitness-NMS~\cite{tychsen2018improving}, Grid R-CNN~\cite{lu2019grid}, TridentNet~\cite{li2019scale}, \textit{etc.} Some other approaches pay more attention to the detection speed. They usually do not have the proposal verification stage. The representative approaches includes SSD~\cite{liu2016ssd}, DSSD~\cite{fu2017dssd}, RON~\cite{kong2017ron}, YOLOv2~\cite{redmon2017yolo9000}, RetinaNet~\cite{lin2018focal}, RefineDet~\cite{zhang2018single}, AlignDet~\cite{chen2019revisiting}, ATSS~\cite{zhang2020bridging}, M2Det~\cite{zhao2019m2det}, GFL~\cite{li2020generalized}, FreeAnchor~\cite{zhang2019freeanchor}, FSAF~\cite{zhu2019feature}, \textit{etc.}

Despite the great success of the application of the anchors, they suffer many drawbacks, \eg, a large number of anchors are often required to ensure a sufficiently high IoU (intersection over union) rate with the ground-truth objects, and the size and aspect ratio of each anchor box need to be manually designed. Therefore, the very neat \textit{anchor-free} proposals are proposed. The anchor-free proposals abandon the anchors and represent objects as the points within objects. The key of the anchor-free proposals is to the accurately predict the labels of the relative sparse points and the distances from the points to the object borders. Typical approaches including YOLO series~\cite{redmon2016you, ge2021yolox}, FCOS~\cite{tian2020fcos}, Objects as Points~\cite{zhou2019objects}, FoveaBox~\cite{kong2020foveabox}, SAPD~\cite{zhu2019soft}, RepPoints series~\cite{yang2019reppoints, chen2020reppoints}, \textit{etc.}

\noindent \textbf{Bottom-up approaches} detects the individual parts of objects instead of perceiving the objects as a whole. Subsequently, the individual parts that belongs to the same object will be grouped together by some trainable post-process algorithms. The bottom-up approach dates back to the pre deep learning era, Felzenszwalb~\textit{et al.} represent objects using mixtures of multi-scale deformable part models which know as DPM~\cite{felzenszwalb2009object}. 
Recently, the keypoint estimation~\cite{newell2017associative} inspires the object detection to recognize the objects by detecting and grouping the keypoints, \eg, CornerNet~\cite{law2018cornernet} and CornerNet-lite~\cite{law2019cornernet} detects objects as paired corners, while ExtremeNet~\cite{zhou2019bottom} detects four extreme points (top-most, left-most, bottom-most, right-most) of an object. Since the bottom-up approaches do not need the anchors, they belong to a kind of the anchor-free detectors. Most of the Bottom-up approaches are based on a state-of-the-art keypoint estimation framework~\cite{cao2017realtime, xiao2018simple}, which also brings some drawbacks, \eg, they rely too much on the heatmap with high-resolution, the inference usually too slow, \textit{etc.}

\section{Our Approach}\label{sec:approach}
\subsection{Baseline and Motivation}\label{subsec:baseline}
We notice that there are two important choices of object detection, top-down or bottom-up? Based on the discussions in the above sections, we infer that \textit{bottom-up approaches have better potential in locating objects with arbitrary geometry, and thus a higher recall}. This is because most top-down approaches work by the anchors, which is very empirical (\eg, to improve efficiency, only the anchors with common sizes and aspect ratios are considered) and their shapes and locations are relatively fixed, although the subsequent bounding-box regression process could slightly change their states. Therefore, the detectors tend to miss the object with a peculiar shape. Fig.~\ref{fig:faster_result} has shown a typical example. We also provide a quantitative study, shown in Table~\ref{tab:false_negatives}. Three representative approaches as well as our work are evaluated on the MS-COCO validation dataset. Table~\ref{tab:false_negatives} shows that the top-down approaches
report significantly lower recalls than the bottom-up approaches, especially for the object with peculiar geometry, \eg, with the scale that larger than $300^2$ pixels or aspect ratio that larger than $5:1$. This is no surprise because, on the one hand, for Faster R-CNN~\cite{ren2016faster}, a typical anchor-based top-down approach, there are no pre-defined anchors could match these objects. On the other hand, for FCOS~\cite{tian2020fcos}, a typical anchor-free top-down approach, it is difficult to accurately regress the long distances from the border to the proposal. Since the bottom-up approaches usually detects the individual parts of objects and group them into the objects, they somewhat avoid this problem. Moreover, we report the results of the proposed CenterNet, which demonstrates that CenterNet inherits the merits of the bottom-up approaches for locating objects flexibly, especially with peculiar geometries.


\begin{table}[t]
\centering
\caption{Average false discovery (AF) rate ($\%$) of CornerNet. AF reflects the distribution of incorrect bounding boxes (false positives). The results suggest the false positives account for a large proportion.}
\resizebox{.48\textwidth}{!}{
\begin{tabular}{|l|ccccccc|}
\hline
Method & AF & AF$_{5}$ & AF$_{25}$ & AF$_{50}$ & AF$_{\mathrm{S}}$ & AF$_{\mathrm{M}}$ & AF$_{\mathrm{L}}$\\
\hline
CornerNet & 37.8 & 32.7 & 36.8 & 43.8 & 60.3 & 33.2 & 25.1 \\
\hline
\end{tabular}}
\label{tab:af}
\end{table}

\begin{figure*}[!tb]
  \centering 
  \includegraphics[width=1\textwidth]{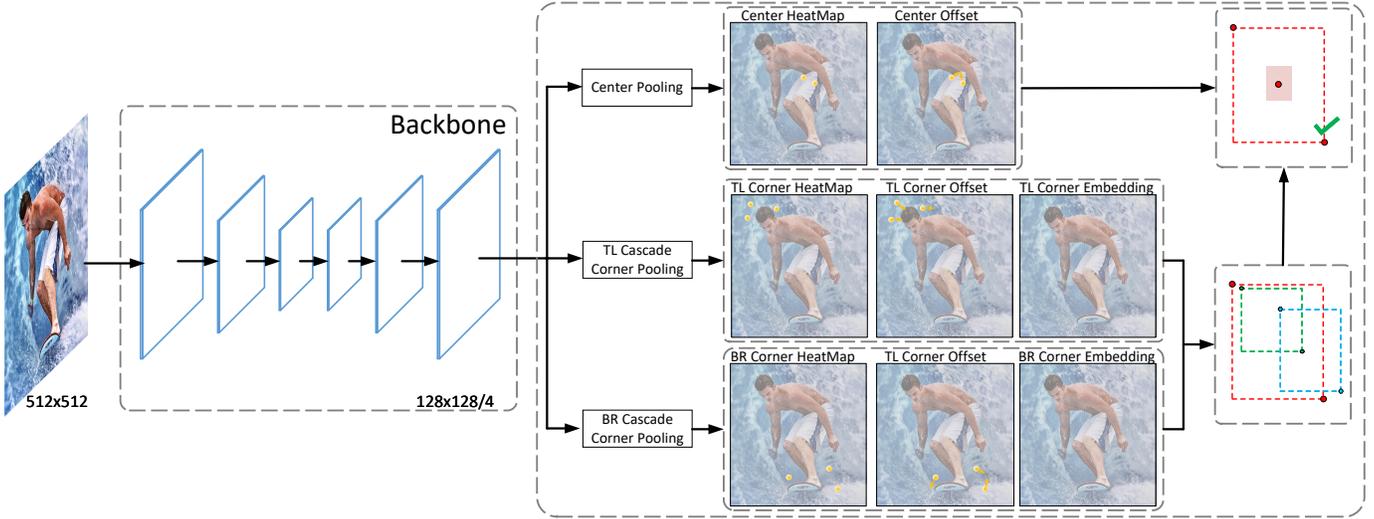}
  \caption{Single-resolution detection framework of CenterNet. A convolutional backbone network applies cascade corner pooling and center pooling to output two corner heatmaps and a center keypoint heatmap, respectively. Note that the heatmaps are multi-class heatmaps, which means that the number of channel of each heatmap equals to the number of the classes in the dataset. Similar to CornerNet, a pair of detected corners and the similar embeddings are used to detect a potential bounding boxes. Then the detected center keypoints are used to determine the final bounding boxes.} 
  \label{fig:hg_centernet} 
\end{figure*}

Although the bottom-up approaches enjoy a high recall, they often generate many false positives. Take CornerNet~\cite{law2018cornernet} as an example, it produces two heatmaps for detecting corners: a heatmap of top-left corners and a heatmap of bottom-right corners. The heatmaps represent the locations of keypoints of different categories and assigns a confidence score for each keypoint. Besides, it also predicts an embedding and a group of offsets for each corner (as shown in Fig~\ref{fig:hg_centernet}). The embeddings are used to identify if two corners are from the same object. The offsets learn to remap the corners from the heatmaps to the input image. For generating object bounding boxes, top-$k$ left-top corners and bottom-right corners are selected from the heatmaps according to their scores, respectively. Then, the distance of the embedding vectors of a pair of corners is calculated to determine if the paired corners belong to the same object. An object bounding box is generated if the distance is less than a threshold. The bounding box is assigned a confidence score, which equals to the average scores of the corner pair. 

In Table~\ref{tab:af}, we provide a deeper analysis of CornerNet. We count the AF\footnote{$\mathrm{AF}=1-\mathrm{AP}$, where AP denotes the average precision at $\mathrm{IoU = [0.05 : 0.05 : 0.5]}$ on the MS-COCO dataset. Also, $\mathrm{AF}_{i} = 1 - \mathrm{AP}_{i}$, where $\mathrm{AP_i}$ denotes the average precision at $\mathrm{IoU} = i/100$, $\mathrm{AF_{scale}} = 1 - \mathrm{AP_{scale}}$, where $\mathrm{scale} = \left\{\mathrm{small, medium, large}\right\}$, denotes the scale of object.} (average false discovery) rate of CornerNet on the MS-COCO validation dataset, defined as the proportion of the incorrect bounding boxes. The quantitative results demonstrate the incorrect bounding boxes account for a large proportion even at low IoU thresholds, \eg, CornerNet obtains $32.7\%$ AF at $\mathrm{IoU = 0.05}$. This means in average, $32.7$  out of every $100$ object bounding boxes have IoU lower than $0.05$ with the ground-truth. The small incorrect bounding boxes are even more, which achieves $60.3\%$ AF. One of the possible reasons lies in that CornerNet cannot look into the regions inside the bounding boxes. To make CornerNet~\cite{law2018cornernet} perceive the visual patterns in bounding boxes, one potential solution is to adapt CornerNet into a two-stage detector, which uses the RoI pooling~\cite{girshick2015fast} to look into the visual patterns in bounding boxes. However, it is known that such a paradigm is computationally expensive.

In this paper, we propose a highly efficient alternative called \textbf{CenterNet} to explore the visual patterns within each bounding box. For detecting an object, our approach uses a triplet, rather than a pair, of keypoints. By doing so, our approach only pays attention to the center information, the cost of our approach is minimal, but partially inherits the functionality of RoI pooling. Furthermore, we design two frameworks, which detect objects on a single-resolution feature map and on multi-resolution feature maps, respectively. The former is applied to the keypoint estimation network, which we hope to detect the corners and the center keypoints better. While the later is more popular in the object detection due the better generality and richer detection receptive fields. The two frameworks are slightly different in design details, we will give a detailed introduction in the next subsection. 

\subsection{Object Detection as Keypoint Triplets}\label{triplets}
\subsubsection{Single-resolution detection framework}\label{single_frame}
Inspired by the pose estimation, we apply the networks that commonly used in pose estimation to better detect the corners and center keypoints, most of which detect the keypoints on a single-resolution feature map, \eg, the hourglass network~\cite{newell2016stacked}. The overall network architecture is shown in Figure~\ref{fig:hg_centernet}. We represent each object by a center keypoint and a pair of corners. Specifically, we embed a heatmap for the center keypoints on the basis of CornerNet and predict the offsets of the center keypoints. Then, we use the method proposed in CornerNet~\cite{law2018cornernet} to generate top-$k$ bounding boxes. However, to effectively filter out the incorrect bounding boxes, we leverage the detected center keypoints and resort to the following procedure: (1) select top-$k$ center keypoints according to their scores; (2) use the corresponding offsets to remap these center keypoints to the input image; (3) define a central region for each bounding box and check if the central region contains center keypoints. Note that the class labels of the checked center keypoints should be same as that of the bounding box; (4) if a center keypoint is detected in the central region, we will preserve the bounding box. The score of the bounding box will be replaced by the average scores of the three points,~\ie,~the top-left corner, the bottom-right corner and the center keypoint. If there are no center keypoints detected in its central region, the bounding box will be removed.

\begin{figure*}[!tb]
  \centering 
  \includegraphics[width=1\textwidth]{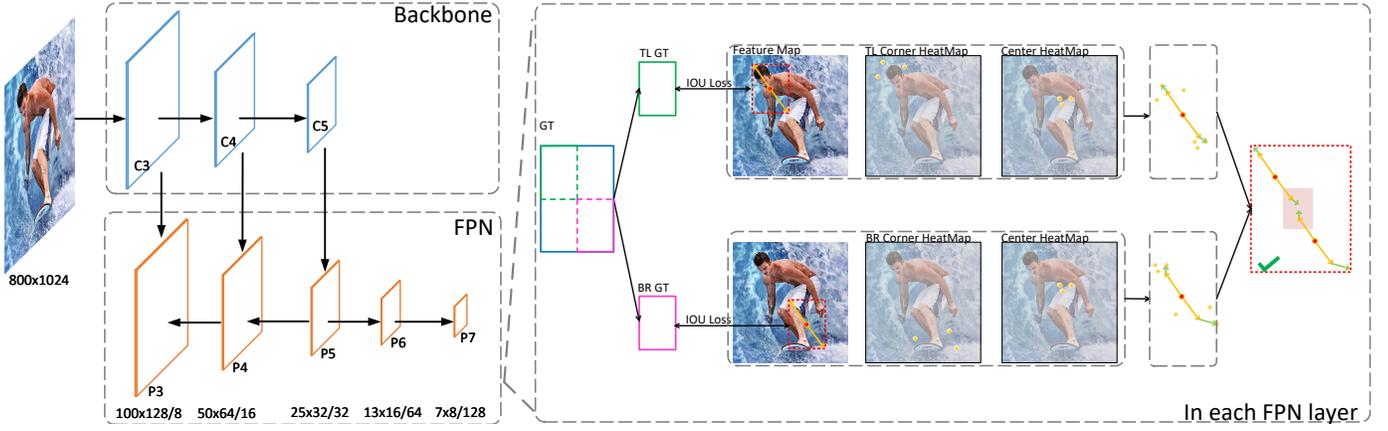}
  \caption{Multi-resolution detection framework of CenterNet. A convolutional backbone network outputs three feature maps, which are C3–C5, to connect a feature pyramid structure (FPN). Then FPN outputs P3–P7 feature maps as the final prediction layers. In each prediction layer, we use heatmap and regression to predict the keypoints, respectively. In the heatmap-based prediction, we predict three light binary heatmaps for predicting corners and center keypoints. In the regression-based prediction, to decouple the top-left and bottom-right corners, we divide the ground-truth boxes into four sub-ground-truth boxes along the geometric center, and we select the top-left and bottom-right sub-ground-truth boxes to supervise the regression, respectively. During the inference, the regressed vectors act as a cue to find the closest keypoints on the corresponding heatmaps to refine the locations of the keypoints. Finally, the predicted keypoint triplets are used to determine the final bounding boxes.} 
  \label{fig:py_centernet} 
\end{figure*}

\subsubsection{Multi-resolution detection framework}\label{multi_frame}
The overall network architecture is shown in Figure~\ref{fig:py_centernet}. It starts with a backbone (\eg, the ResNet~\cite{he2016deep}, ResNeXt~\cite{xie2017aggregated}, \textit{etc.}) that extracts features from the input image. We select C3–C5 feature maps from the backbone as the input of a feature pyramid structure (FPN). Then the FPN outputs P3–P7 feature maps as the final prediction layers. In each prediction layer, we use heatmap and regression to predict the keypoints, respectively. In the heatmap-based prediction, we predict three light binary heatmaps for predicting corners and center keypoints. The resolution of the heatmap is the same with the prediction layer, therefore, we predict a extra offset for each keypoint to learn to remap the keypoint from the heatmap to the input image. In the regression-based prediction, to decouple the top-left and bottom-right corners, we divide the ground-truth boxes into four sub-ground-truth boxes along the geometric center, and we select the top-left and bottom-right sub-ground-truth boxes to supervise the regression, respectively. Take the regression of the top-left box as a example, we select some feature points within the top-left sub-ground-truth boxes, each selected feature point predicts two vectors, which point to the top-left corners and center keypoint. Moreover, we also assign each selected feature point a class label to supervise the classification. We apply the common anchor-free detection methods to train the network to predict the sub-bounding boxes (such as FCOS~\cite{tian2020fcos} and RepPoints~\cite{chen2020reppoints}). As a side comment, here we emphasize that the regression accuracy of sub-bounding boxes will be higher than that of complete bounding boxes, because Table~\ref{tab:false_negatives} shows that anchor-free methods like FCOS~\cite{tian2020fcos} suffer from the low accuracy of the long distance regression, while our the sub-bounding boxes effectively halve the regression distances. 

During the inference, the regressed vectors act as a cue to find the closest keypoints on the corresponding heatmaps to refine the locations of the keypoints. Next, each valid pair of keypoints defines a bounding box. Here by valid we mean that two keypoints belong to the same class (\ie, the corresponding top-left and bottom-right sub-bounding box of the same class), and the $x$ and $y$ coordinates of the top-left point are smaller than that of the bottom-right point, respectively. Finally, we define a central region for each bounding box and check if the central region contains both the predicted center keypoints. If there is at most one center keypoint detected in its central region, the bounding box will be removed. The score of the bounding box will be replaced by the average scores of the the points,~\ie, the top-left corner, the bottom-right corner and the center keypoints. 

\subsubsection{Central region definition}\label{central_region}

\begin{figure}[h]
  \centering 
  \subfigure[]{ 
    \includegraphics[width = 0.16\textwidth]{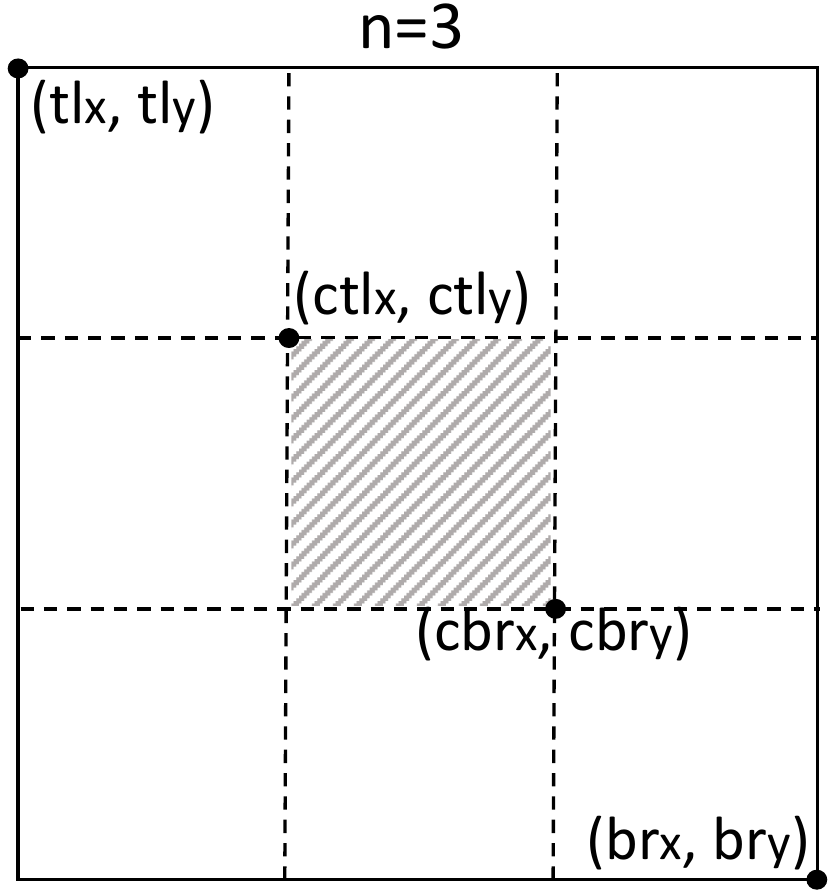}
    \label{fig:ncs1} 
  } 
  \hspace{3ex}
  \subfigure[]{ 
    \includegraphics[width = 0.16\textwidth]{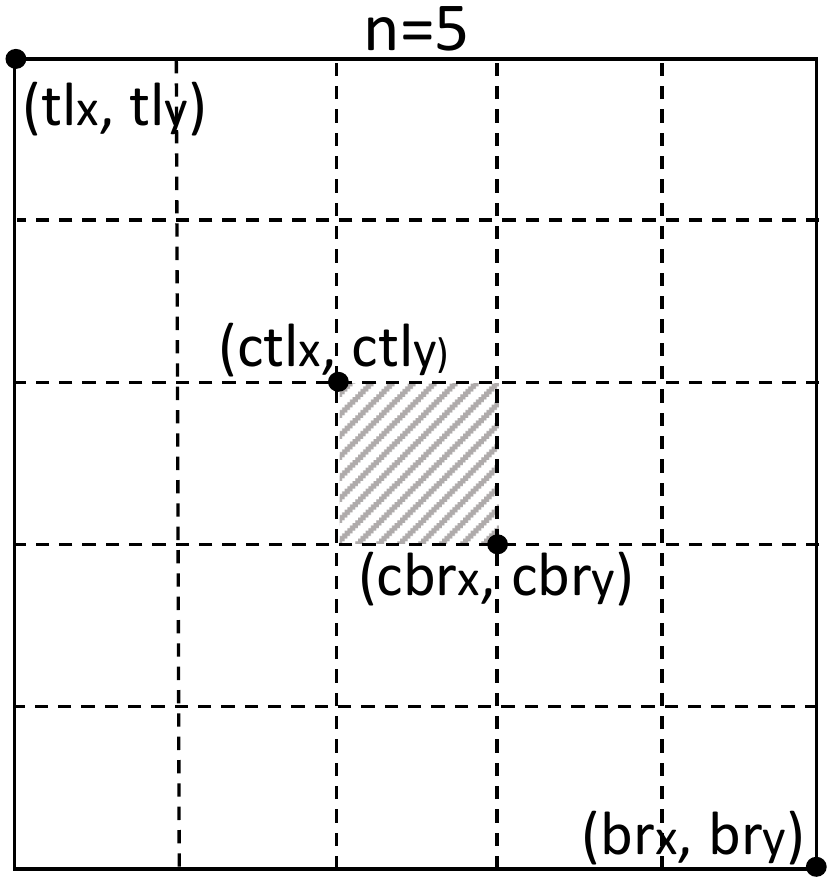}
    \label{fig:ncs2} 
  }
  \caption{(a) The central region when $n=3$. (b) The central region when $n=5$. The solid rectangle and the shaded region denote the predicted bounding box and the scalable central region, respectively.}
  \label{fig:ncs}
\end{figure}

The size of the central region in the bounding box affects the detection results. For example, small central regions lead to a low recall rate for small bounding boxes, while large central regions lead to a low precision for large bounding boxes. Therefore, we propose a scale-aware central region to adaptively fit the size of bounding boxes. The scale-aware central region tends to generate a relatively large central region for a small bounding box and a relatively small central region for a large bounding box. Let $\mathrm{tl_{x}}$ and $\mathrm{tl_{y}}$ denote the coordinates of the top-left corner of $i$ and $\mathrm{br_{x}}$ and $\mathrm{br_{y}}$ denote the coordinates of the bottom-right corner of $i$. 
Define a central region $j$. Let $\mathrm{ctl_{x}}$ and $\mathrm{ctl_{y}}$ denote the coordinates of the top-left corner of $j$ and $\mathrm{cbr_{x}}$ and $\mathrm{cbr_{y}}$ denote the coordinates of the bottom-right corner of $j$. Then $\mathrm{tl_{x}}$, $\mathrm{tl_{y}}$, $\mathrm{br_{x}}$, $\mathrm{br_{y}}$, $\mathrm{ctl_{x}}$, $\mathrm{ctl_{y}}$, $\mathrm{cbr_{x}}$ and $\mathrm{cbr_{y}}$ should satisfy the following relationship:
\begin{equation} \label{eq:ncs1}
\small
\left\{
\begin{aligned}
&\mathrm{ctl_{x}} = \frac{(n+1)\mathrm{tl_{x}}+(n-1)\mathrm{br_{x}}}{2n} \\
&\mathrm{ctl_{y}} = \frac{(n+1)\mathrm{tl_{y}}+(n-1)\mathrm{br_{y}}}{2n} \\
&\mathrm{cbr_{x}} = \frac{(n-1)\mathrm{tl_{x}}+(n+1)\mathrm{br_{x}}}{2n} \\
&\mathrm{cbr_{y}} = \frac{(n-1)\mathrm{tl_{y}}+(n+1)\mathrm{br_{y}}}{2n} \\
\end{aligned}
\right.
\end{equation}
where $n$ is odd and determines the scale of the central region $j$. In this paper, $n$ is set to be $3$ and $5$ for the scales of bounding boxes less than and greater than $150$, respectively. Figure~\ref{fig:ncs} shows two central regions when $n=3$ and $n=5$, respectively. According to Equation~(\ref{eq:ncs1}), we could determine a scale-aware central region and then check whether the central region contains center keypoints.

\subsection{Enriching Center and Corner Information}\label{Enriching}

\begin{figure}[tb]
  \centering 
  \subfigure[]{ 
    \includegraphics[width=0.11\textheight]{figures/CenterPooling.pdf}
    \label{fig:ct}
  } 
  \hspace{-1.5ex}
  \subfigure[]{ 
    \includegraphics[width=0.11\textheight]{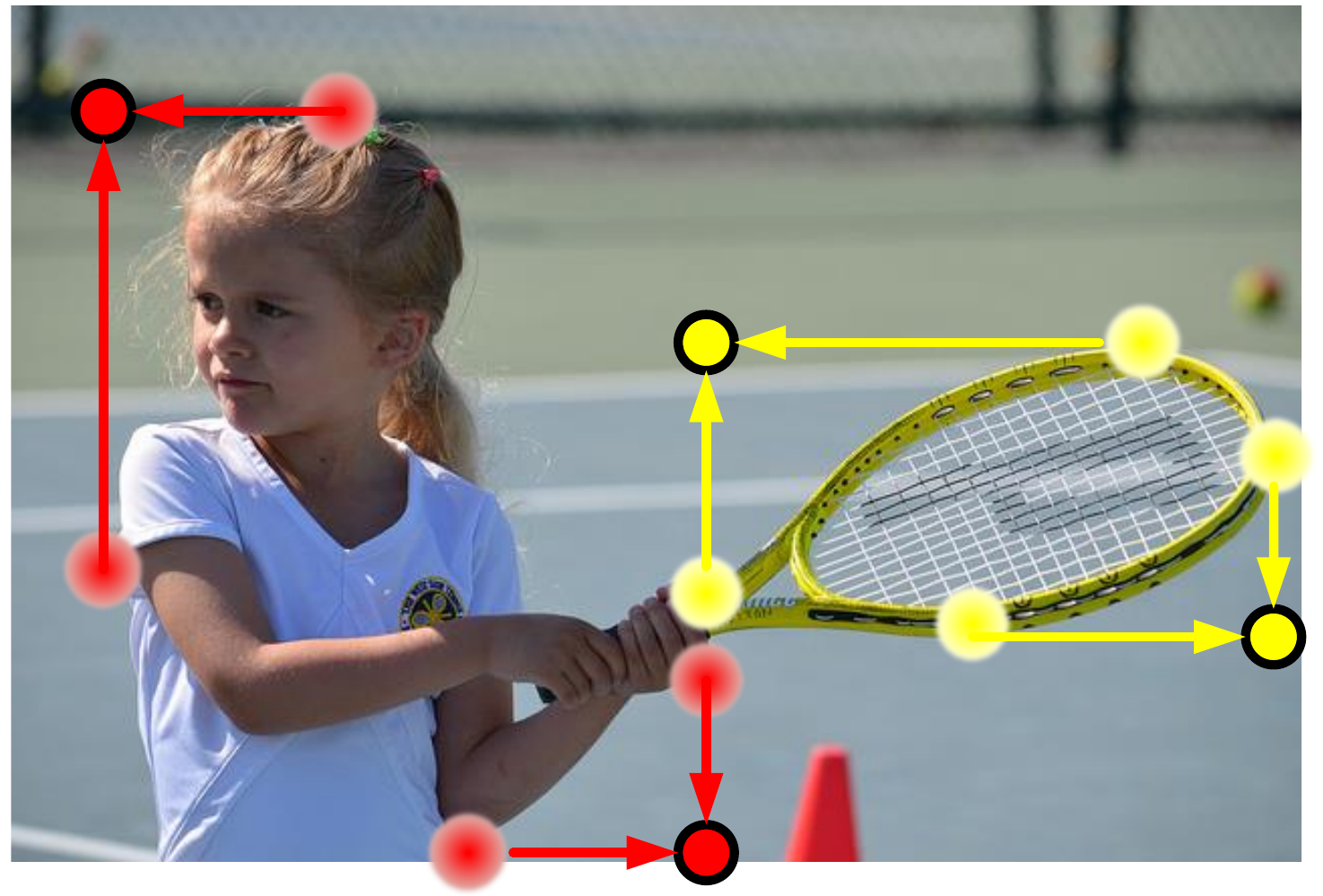}
    \label{fig:cp} 
  } 
  \hspace{-1.5ex}
  \subfigure[]{ 
    \includegraphics[width=0.11\textheight]{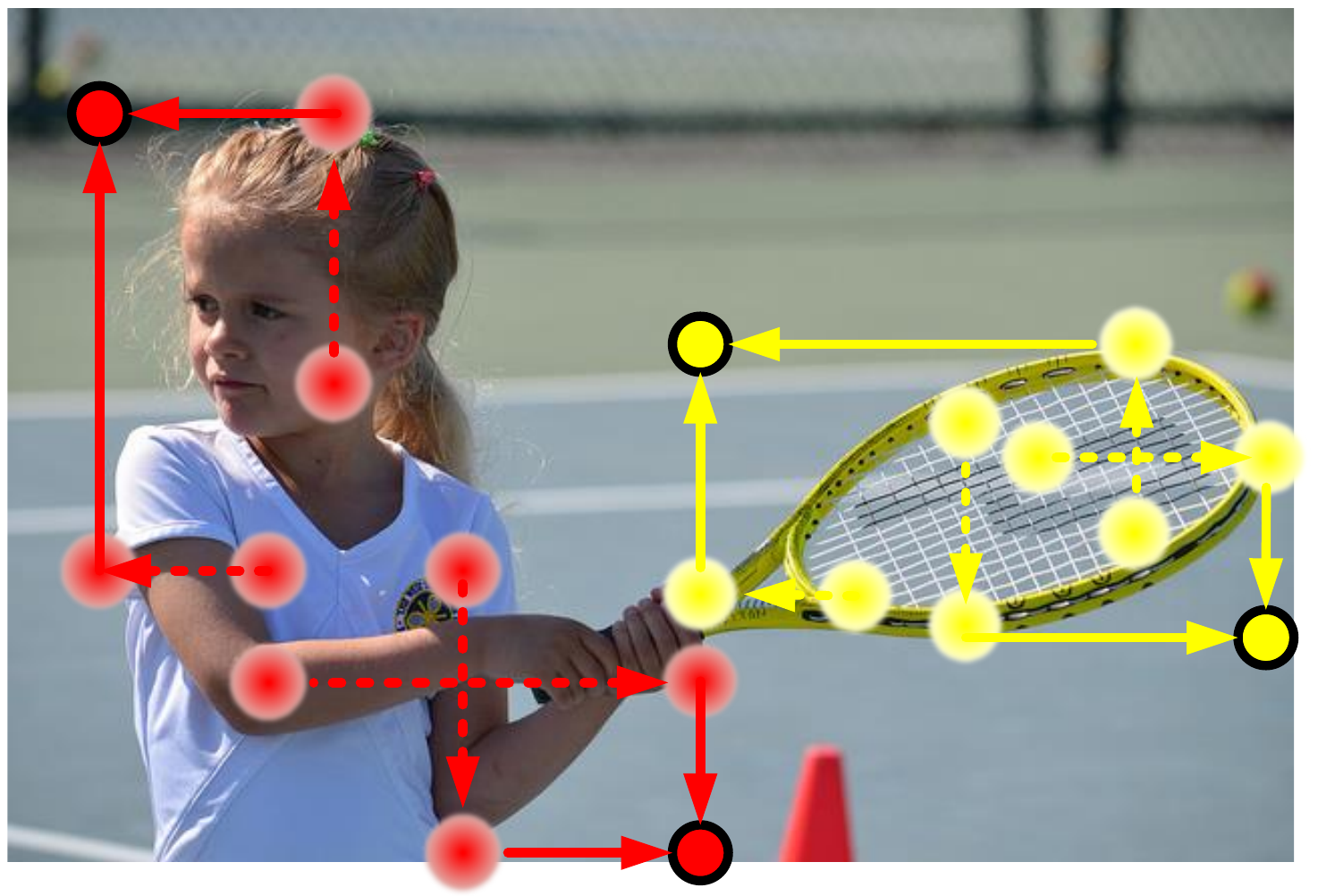}
    \label{fig:ccp} 
  } 
  \caption{(a) Center pooling takes the maximum values in both horizontal and vertical directions. (b) Corner pooling only takes the maximum values in boundary directions. (c) Cascade corner pooling takes the maximum values in both boundary directions and internal directions of objects.} 
  \label{fig:cornerpooling}
\end{figure}

Both center keypoints and corners have rigorous geometric relationships with the objects but contains limited visual patterns of objects. We train the network in a fully supervised way to learn the geometric relationships and the limited visual features so as to locate keypoints. If we introduce more visual patterns for both center keypoints and corners, they will be better detected.

\noindent\textbf{Center pooling.}~The geometric centers of objects do not always convey very recognizable visual patterns (\eg, the human head contains strong visual patterns, but the center keypoint is often in the middle of the human body). To address this issue, we propose center pooling to capture richer and more recognizable visual patterns. Figure~\ref{fig:ct} shows the principle of center pooling. The detailed process of center pooling is as follows: the backbone outputs a feature map and to determine whether a pixel in the feature map is a center keypoint, we need to find the maximum value in both the horizontal and vertical directions and add these values together. By doing so, center pooling helps improve the detection of center keypoints.

\noindent\textbf{Cascade corner pooling.}~Corners are often outside objects, which lack local appearance features. CornerNet~\cite{law2018cornernet} uses corner pooling to address this issue. The principle of corner pooling is shown in Figure~\ref{fig:cp}. Corner pooling aims to find the maximum values on the boundary directions to determine corners. However, this makes corners sensitive to edges. To address this problem, we need to enable corners to extract features from central regions of the object. The principle of cascade corner pooling is presented in Figure~\ref{fig:ccp}. Cascade corner pooling first looks along a boundary to find a maximum boundary value and then looks inside the box along with the location of the boundary maximum value\footnote{\scriptsize For the topmost, leftmost, bottommost and rightmost boundary, look vertically towards the bottom, horizontally towards the right, vertically towards the top and horizontally towards the left, respectively.} to find an internal maximum value; finally, the two maximum values are added together. By cascade corner pooling, the corners obtain both the boundary information and the visual patterns of objects.

\begin{figure}[tb]
  \centering 
  \includegraphics[width=0.48\textwidth]{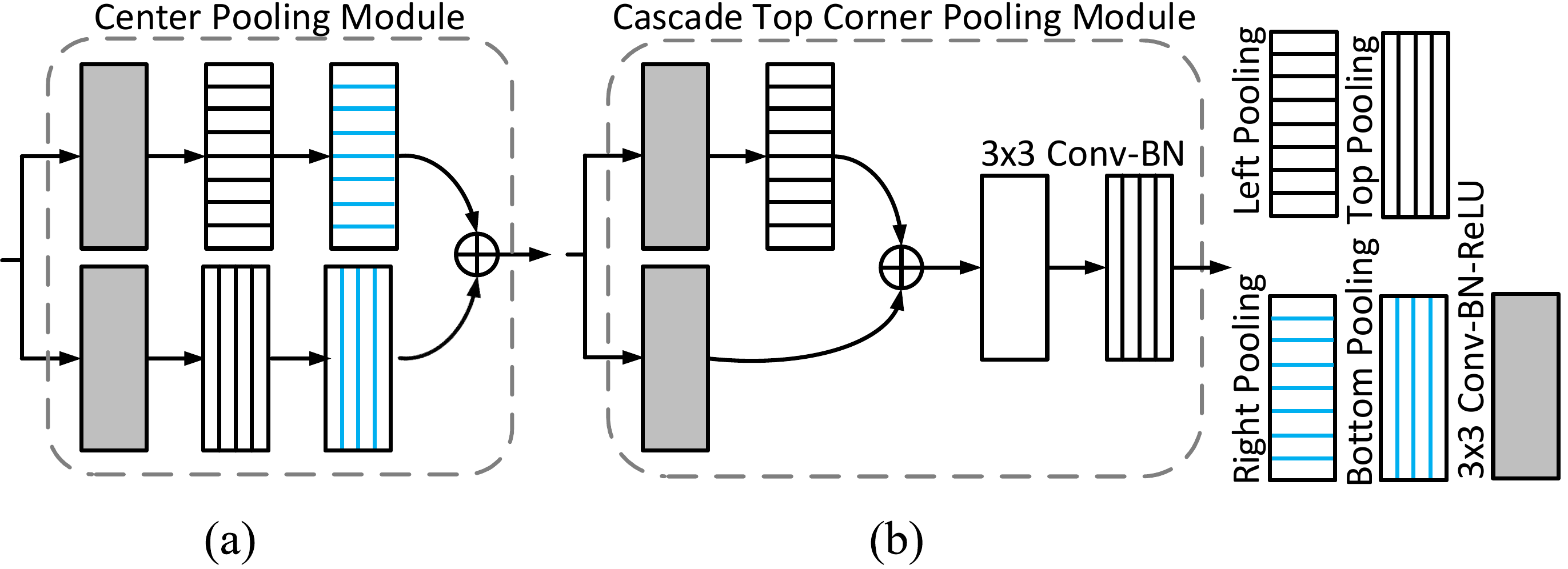}
  \caption{The structures of the center pooling module (a) and the cascade top corner pooling module (b). We achieve center pooling and cascade corner pooling by combining corner pooling in different directions.} 
  \label{fig:PoolingStructure} 
\end{figure}

Both center pooling and the cascade corner pooling could be easily achieved by applying the corner pooling~\cite{law2018cornernet} in different directions. Figure~\ref{fig:PoolingStructure}{\color{red}(a)} shows the structure of the center pooling module. To take a maximum value in a specific direction, \eg, the horizontal direction, we only need to connect the left pooling and the right pooling in sequence. Figure~\ref{fig:PoolingStructure}{\color{red}(b)} shows the structure of a cascade top corner pooling module, in which the white rectangle denotes a $3\times3$ convolution followed by batch normalization. Compared with the top corner pooling in CornerNet~\cite{law2018cornernet}, a left corner pooling is added before the top corner pooling.

\subsection{Training and Inference }\label{Training}
\textbf{Training.} We train CenterNet on $8$ Tesla V100 (32GB) GPUs. For single-resolution detection framework, our direct baseline is CornerNet~\cite{law2018cornernet}. Following it, we use the stacked hourglass network (Hourglass)~\cite{newell2016stacked} with $52$ and $104$ layers as the backbone -- the latter has two hourglass modules while the former has only one. All modifications on the hourglass architecture, made by~\cite{law2018cornernet}, are preserved. The network is trained from scratch when we use the Hourglass as the backbone. In addition, to show that the framework generalizes to other network architectures, we investigate another backbone named HRNet~\cite{sun2019deep,sun2019high}, which enjoys the ability to maintain high-resolution representations throughout feature extraction. The resolution of the input image is $511\times511$, leading to heatmaps of the size $128\times128$. We use the data augmentation strategy presented in~\cite{law2018cornernet} to train a robust model. Adam~\cite{kingma2014adam} is used to optimize the training loss:
\begin{equation} \label{loss_hg} 
\small
\mathrm{L_s = L_{kp}^\mathrm{co} + L_{kp}^\mathrm{ce} + \alpha L_{pull}^\mathrm{co} + \beta L_{push}^\mathrm{co} + \gamma\left( L_{off}^\mathrm{co} + L_{off}^\mathrm{ce} \right)},
\end{equation}
where $\mathrm{L_{kp}^\mathrm{co}}$ and $\mathrm{L_{kp}^\mathrm{ce}}$ denote the focal losses, which are used to train the network to detect corners and center keypoints, respectively. $\mathrm{L_{pull}^\mathrm{co}}$ is a ``pull'' loss for corners, which is used to minimize the distance of the embedding vectors that belongs to the same objects. $\mathrm{L_{push}^\mathrm{co}}$ is a  ``push'' loss for corners that is used to maximize the distance of the embedding vectors that belong to different objects. $\mathrm{L_{off}^\mathrm{co}}$ and $\mathrm{L_{off}^\mathrm{ce}}$ are $\ell_1$-losses~\cite{girshick2015fast}, which are used to train the network to predict the offsets of corners and center keypoints, respectively. $\mathrm{\alpha}$, $\mathrm{\beta}$ and $\mathrm{\gamma}$ denote the weights for corresponding losses and are set to $0.1$, $0.1$ and $1$, respectively. $\mathrm{L_{kp}}$, $\mathrm{L_{pull}}$, $\mathrm{L_{push}}$ and $\mathrm{L_{off}}$ are all defined in CornerNet, and we suggest referring to~\cite{law2018cornernet} for details. We use a batch size of $48$. The maximum number training epochs is $100$. We use a learning rate of $2.5\times10^{-4}$ for the first $88$ epochs and then continue training $12$ epochs with a rate of $2.5\times10^{-5}$.  

For multi-resolution detection framework, we use ResNet~\cite{he2016deep}, Res2Net~\cite{gao2019res2net}, ResNeXt~\cite{xie2017aggregated} and Swin-Transformer~\cite{liu2021swin} with the weights pre-trained on ImageNet~\cite{deng2009imagenet} as our backbones, respectively. The
FPN~\cite{lin2017feature} structure is applied to output the detection layers with different scales. Both single-scale and multi-scale training strategies are applied. For the single-scale training, the shorter side of each input image is $800$, while for the multi-scale training, the shorter side of each input image is randomly selected from a range of $[480, 960]$. We use the data augmentation strategy presented in~\cite{yang2019reppoints} to train a robust model. Stochastic gradient descent (SGD) is used to optimize the training loss:
\begin{equation} \label{loss_py} 
\small
\begin{aligned}
\mathrm{L_m = \frac{1}{2}\left(L_{cls}^\mathrm{tl} + L_{cls}^\mathrm{br}\right) + \frac{\hat{\alpha}}{2}\left(L_{reg}^\mathrm{tl} + L_{reg}^\mathrm{br}\right)} \\
\mathrm{+ \hat{\beta}\left(L_{kp}^\mathrm{co} + L_{kp}^\mathrm{ce}\right) + \hat{\gamma}\left( L_{off}^\mathrm{co} + L_{off}^\mathrm{ce} \right)},
\end{aligned}
\end{equation}
where $\mathrm{L_{cls}^\mathrm{br}}$ and $\mathrm{L_{cls}^\mathrm{br}}$ denote the focal losses, which are used to train the network to classify the the top-left and bottom-right sub-bounding boxes, respectively. $\mathrm{L_{reg}^\mathrm{tl}}$ and $\mathrm{L_{reg}^\mathrm{br}}$ denote the GIoU loss~\cite{rezatofighi2019generalized}, which are used to train the network to regress the the top-left and bottom-right sub-bounding boxes, respectively. $\mathrm{\hat{\alpha}}$, $\mathrm{\hat{\beta}}$ and $\mathrm{\hat{\gamma}}$ denote the weights for corresponding losses and are set to $2$, $0.25$ and $1.0$, respectively. We use a batch size of $16$. The maximum number training epochs is $24$. We use a learning rate of $0.01$ for the first $16$ epochs and then the learning rate decays by a factor of $10$ after the $16$th epoch and the $22$th epoch, respectively.  

\noindent\textbf{Inference.} For single-resolution detection framework, we following~\cite{law2018cornernet}. For the single-scale testing, we input both the original and horizontally flipped images with the original resolutions into the network. For multi-scale testing, we input both the original and horizontally flipped images with resolutions of $0.6, 1, 1.2, 1.5$ and $1.8$. We select top $70$ center keypoints, top $70$ top-left corners and top $70$ bottom-right corners from the heatmaps to detect the bounding boxes. We flip the bounding boxes detected in the horizontally flipped images and mix them into the original bounding boxes. Soft-NMS~\cite{bodla2017soft} is used to remove the redundant bounding boxes. We finally select the top $100$ bounding boxes according to their scores as the final detection results.

For multi-resolution detection framework, we follow~\cite{yang2019reppoints}. For the single-scale testing, we resize each image to a shorter side of $800$ to input the network, while for the multi-scale testing, we first resize each image to a shorter side of $[400, 600, 800, 1000, 1200, 1400]$, then the detection results from all scales are merged. A NMS with a threshold of $0.6$ is applied to remove the redundant results. Flip argumentation is added to only the multi-scale evaluation.

\subsection{Relationship to Prior Works}\label{relationship}
Our approach inherits advantages of bottom-up and top-down approaches. Top-down approaches have the ability of perceiving the global visual contents within proposals, but they usually suffer from the low location accuracies especially for the peculiar shapes. Bottom-up approaches enjoy the ability in locating objects with arbitrary geometries, but often exist many incorrect bounding boxes (false positives). Our approach uses a triplet of keypoints to represent an object, which still is a bottom-up approach but could perceive visual contents within proposals and the cost of is minimal.

\section{Experiments}\label{sec:experiments}
\subsection{Dataset, Metrics and Baseline}\label{sec:setting}
We evaluate our method on the MS-COCO dataset~\cite{lin2014microsoft}. This dataset contains $80$ categories and more than $1.5$ million object instances. A large number of small objects makes it a very challenging dataset. We use the `train2017' set (\ie, $110\mathrm{K}$ training images) for training and testing the results on the test-dev set. We use `val2017' set as the validation set to perform ablation studies and visualization experiments.

The MS-COCO dataset~\cite{lin2014microsoft} uses AP and AR metrics to characterize the performance of a detector. AP represents the average precision rate, which is computed over ten different IoU thresholds (\ie, $0.5:0.05:0.95$) and all categories. AR represents the maximum recall rate, which is computed over a fixed number of detections (\ie, $1$, $10$ and $100$ ) per image and averaged over all categories and the ten different IoU thresholds. Additionally, AP and AR can be used to evaluate the performance under different object scales, including small objects ($\mathrm{area}<32^{2}$), medium objects ($32^{2}<\mathrm{area}<96^{2}$) and large objects ($\mathrm{area}>96^{2}$). AP is considered the single most important metric on the MS-COCO dataset.

\subsection{Comparisons with State-of-the-art Detectors}

\begin{table*}[tb]
\centering
\caption{Performance comparison ($\%$) with the state-of-the-art methods on the MS-COCO test-dev dataset. CenterNet outperforms all existing bottom-up detectors by a large margin and ranks among the top of state-of-the-art top-down detectors. The abbreviations are: `R' -- ResNet~\cite{he2016deep}, `X' -- ResNeXt~\cite{xie2017aggregated}, `HG' -- Hourglass network~\cite{newell2016stacked}, `R2' -- Res2Net~\cite{gao2019res2net}, `$\dagger$' -- multi-scale testing, `SR' -- detecting objects on the single-resolution feature map, `MR' -- detecting objects on the multi-resolution feature maps.}
\resizebox{1.0\textwidth}{!}{
\begin{tabular}{|l|l|cccccc|cccccc|}
\hline
Method & Backbone & AP & AP$_{50}$ & AP$_{75}$ & AP$_\mathrm{S}$ & AP$_\mathrm{M}$ & AP$_\mathrm{L}$ & AR$_1$ & AR$_{10}$ & AR$_{100}$ & AR$_\mathrm{S}$ & AR$_\mathrm{M}$ & AR$_\mathrm{L}$\\
\hline
\hline
\textbf{Top-down:} & & & & & & & & & & & & &\\
YOLOv2~\cite{redmon2017yolo9000} & DarkNet-19 & 21.6 & 44.0 & 19.2 & 5.0 & 22.4 & 35.5 & 20.7 & 31.6 & 33.3 & 9.8 & 36.5 & 54.4 \\
DSOD300~\cite{shen2017dsod} & DS/64-192-48-1 & 29.3 & 47.3 & 30.6 & 9.4 & 31.5 & 47.0 & 27.3 & 40.7 & 43.0 & 16.7 & 47.1 & 65.0 \\
GRP-DSOD320~\cite{shen2017learning} & DS/64-192-48-1 & 30.0 & 47.9 & 31.8 & 10.9 & 33.6 & 46.3 & 28.0 & 42.1 & 44.5 & 18.8 & 49.1 & 65.0 \\
SSD513~\cite{liu2016ssd} & R-101 & 31.2 & 50.4 & 33.3 & 10.2 & 34.5 & 49.8 & 28.3 & 42.1 & 44.4 & 17.6 & 49.2 & 65.8 \\
DSSD513~\cite{fu2017dssd} & R-101 & 33.2 & 53.3 & 35.2 & 13.0 & 35.4 & 51.1 & 28.9 & 43.5 & 46.2 & 21.8 & 49.1 & 66.4 \\
Faster R-CNN~\cite{shrivastava2016beyond} & R-101 & 36.8 & 57.7 & 39.2 & 16.2 & 39.8 & 52.1 & 31.6 & 49.3 & 51.9 & 28.1 & 56.6 & 71.1 \\
RetinaNet~\cite{lin2018focal} & R-101 & 39.1 & 59.1 & 42.3 & 21.8 & 42.7 & 50.2 & - & - & - & - & - & - \\  
Mask R-CNN~\cite{he2018mask} & X-101-64x4d & 39.8 & 62.3 & 43.4 & 22.1 & 43.2 & 51.2 & - & - & - & - & - & - \\
RefineDet~\cite{zhang2018single}~$\dagger$ & R-101 & 41.8 & 62.9 & 45.7 & 25.6 & 45.1 & 54.1 & - & - & - & - & - & -  \\
Cascade R-CNN~\cite{cai2018cascade} & R-101 & 42.8 & 62.1 & 46.3 & 23.7 & 45.5 & 55.2 & - & - & - & - & - & -  \\
FSAF~\cite{zhu2019feature}~$\dagger$ & X-101-64x4d & 44.6 & 65.2 & 48.6 & 29.7 & 47.1 & 54.6& - & - & - & - & - & -  \\
DETR~\cite{carion2020end}& R-101 & 44.9 & 64.7 & 47.7 & 23.7 & 49.5 & 62.3& - & - & - & - & - & -  \\
Objects as Points~\cite{zhou2019objects}~$\dagger$ & HG-104 & 45.1 & 63.9 & 49.3 & 26.6 & 47.1 & 57.7& - & - & - & - & - & -  \\
RepPointsV1~\cite{yang2019reppoints}~$\dagger$ & R-101-DCN & 46.5 & 67.4 & 50.9 & 30.3 & 49.7 & 57.1& - & - & - & - & - & -  \\
FreeAnchor~\cite{zhang2019freeanchor}~$\dagger$&X-101-32x8d &47.3& 66.3 & 51.5 & 30.6 & 50.4 & 59.0& - & - & - & - & - & -  \\
TridentNet~\cite{li2019scale}~$\dagger$ & R-101-DCN & 48.4 & 69.7 & 53.5 & 31.8 & 51.3 & 60.3& - & - & - & - & - & -  \\
FCOS~\cite{tian2020fcos}&X-101-32x8-DCN& 50.4 & 68.9 & 55.0 & 33.2 & 53.0 & 62.7 & - & - & - & - & - & - \\
ATSS~\cite{zhang2020bridging}~$\dagger$ & X-101-64x4d-DCN & 50.7 & 68.9 & 56.3 & 33.2 & 52.9 & 62.4 & - & - & - & - & - & - \\
RepPointsV2~\cite{chen2020reppoints}~$\dagger$ & X-101-64x4d-DCN & 52.1 & 70.1 & 57.5 & 34.5 & 54.6 & 63.6 & - & - & - & - & - & -  \\
GFLV2~\cite{li2021generalized}~$\dagger$ & R2-101-DCN & 53.3 & 70.9 & 59.2 & 35.7 & 56.1 & 65.6 & - & - & - & - & - & -  \\
LSNet~\cite{duan2021location}~$\dagger$&R2-101-DCN&53.5& 71.1 & 59.2 & 35.2 & 56.4 & 65.8 & 39.8 & 66.2 & 71.4 & 53.7 & 74.1 & 85.3  \\
YOLOv4-P7~\cite{wang2020scaled}$\dagger$ & CSP-P7 & 56.0 & 73.3 & 61.2 & 38.9 & 60.0 & 68.6 & - & - & - & - & - & -  \\ 
Swin Transformer~\cite{liu2021swin}$\dagger$ & Swin-L & 58.7 & - & - & - & - & - & - & - & - & - & - & -  \\ 
\hline
\hline
\textbf{Bottom-up:} & & & & & & & & & & & & &\\
DeNet~\cite{tychsen2017denet} & R-101 & 33.8 & 53.4 & 36.1 & 12.3 & 36.1 & 50.8 & 29.6 & 42.6 & 43.5 & 19.2 & 46.9 & 64.3 \\
CornerNet~\cite{law2018cornernet} & HG-52 & 37.8 & 53.7 & 40.1 & 17.0 & 39.0 & 50.5 & 33.9 & 52.3 & 57.0 & 35.0 & 59.3 & 74.7 \\
CornerNet~\cite{law2018cornernet}$\dagger$&HG-52 & 39.4 & 54.9 & 42.3 & 18.9 & 41.2 & 52.7 & 35.0 & 53.5 & 57.7 & 36.1 & 60.1 & 75.1 \\
CornerNet~\cite{law2018cornernet} & HG-104 & 40.5 & 56.5 & 43.1 & 19.4 & 42.7 & 53.9 & 35.3 & 54.3 & 59.1 & 37.4 & 61.9 & 76.9 \\
CornerNet~\cite{law2018cornernet}$\dagger$&HG-104& 42.1 & 57.8 & 45.3 & 20.8 & 44.8 & 56.7 & 36.4 & 55.7 & 60.0 & 38.5 & 62.7 & 77.4 \\
ExtremeNet~\cite{zhou2019bottom}~$\dagger$ & HG-104 & 43.2 & 59.8 & 46.4 & 24.1 & 46.0 & 57.1 & - & - & - & - & - & -\\
CPNDet~\cite{duan2020corner} & HG-52 & 43.9 & 61.6 & 47.5 & 23.9 & 46.3 & 57.1 & 35.1 & 57.6 & 61.9 & 39.5 & 65.5 & 79.5\\
CPNDet~\cite{duan2020corner}~$\dagger$ & HG-52 & 45.8 & 63.9 & 49.7 & 26.8 & 48.4 & 59.4 & 36.6 & 60.3 & 64.4 & 43.5 & 67.9 & 80.9\\
CPNDet~\cite{duan2020corner} & HG-104 & 47.0 & 65.0 & 51.0 & 26.5 & 50.2 & 60.7 & 36.5 & 59.8 & 64.1 & 41.6 & 68.0 & 81.7\\
CentripetalNet~\cite{dong2020centripetalnet}& HG-104 & 48.0 & 65.1 & 51.8 & 29.0 & 50.4 & 59.9 & - & - & - & - & - & -\\
CPNDet~\cite{duan2020corner}~$\dagger$ & HG-104 & 49.2 & 67.3 & 53.7 & 31.0 & 51.9 & 62.4 & 37.9 & 62.9 & 67.2 & 47.9 & 70.5 & 82.4\\
\hline
\textbf{SR-CenterNet} & HG-52 & 41.6 & 59.4 & 44.2 & 22.5 & 43.1 & 54.1 & 34.8 & 55.7 & 60.1 & 38.6 & 63.3 & 76.9 \\
\textbf{SR-CenterNet} & HRNet-W64 & 44.0 & 62.6 & 47.1 & 23.0 & 47.3 & 57.8 & 35.4 & 56.9 & 61.7 & 38.3 & 66.2 & 79.6 \\
\textbf{SR-CenterNet} & HG-104 & 44.9 & 62.4 & 48.1 & 25.6 & 47.4 & 57.4 & 36.1 & 58.4 & 63.3 & 41.3 & 67.1 & 80.2 \\
\textbf{SR-CenterNet}~$\dagger$ & HG-52 & 43.5 & 61.3 & 46.7 & 25.3 & 45.3 & 55.0 & 36.0 & 57.2 & 61.3 & 41.4 & 64.0 & 76.3 \\
\textbf{SR-CenterNet}~$\dagger$ & HRNet-W64 & 46.3 & 64.7 & 49.8 & 26.6 & 49.6 & 59.3 & 36.8 & 58.6 & 62.9 & 42.1 & 66.9 & 79.0 \\
\textbf{SR-CenterNet}~$\dagger$ & HG-104 & 47.0 & 64.5 & 50.7 & 28.9 & 49.9 & 58.9 & 37.5 & 60.3 & 64.8 & 45.1 & 68.3 & 79.7 \\
\textbf{MR-CenterNet} & R-50 & 46.4 & 63.7 & 50.3 & 27.1 & 48.9 & 58.8 & 36.2 & 60.0 & 64.2 & 41.1 & 68.5 & 81.9 \\
\textbf{MR-CenterNet} & R-101 & 47.7 & 65.1 & 51.9 & 27.8 & 50.5 & 60.6 & 37.1 & 61.1 & 65.4 & 41.6 & 70.0 & 83.4 \\
\textbf{MR-CenterNet} & R-101-DCN & 49.8 & 67.3 & 54.1 & 29.1 & 52.6 & 64.2 & 37.8 & 62.0 & 66.3 & 43.6 & 70.8 & 84.0 \\
\textbf{MR-CenterNet} & X-101 & 49.3 & 67.0 & 53.7 & 30.1 & 52.2 & 62.1 & 37.5 & 61.8 & 66.0 & 43.9 & 70.2 & 83.2 \\
\textbf{MR-CenterNet} & X-101-DCN & 50.8 & 68.6 & 55.4 & 30.7 & 53.4 & 65.3 & 38.2 & 62.7 & 66.9 & 44.9 & 71.0 & 84.6 \\
\textbf{MR-CenterNet} & R2-101 & 50.2 & 67.9 & 54.7 & 30.5 & 53.4 & 63.2 & 38.1 & 62.7 & 67.0 & 44.8 & 71.6 & 84.0 \\
\textbf{MR-CenterNet} & R2-101-DCN & 51.5 & 69.2 & 56.2 & 31.0 & 54.4 & 65.7 & 38.5 & 63.1 & 67.5 & 45.6 & 71.7 & 84.6 \\
\textbf{MR-CenterNet}~$\dagger$ & R2-101-DCN & 53.7 & 70.9 & 59.7 & 35.1 & 56.0 & 66.7 & 39.8 & 66.6 & 71.8 & 54.3 & 74.5 & 86.2 \\
\textbf{MR-CenterNet} & Swin-L & 53.2 & 71.4 & 57.4 & 33.2 & 56.2 & 68.7 & 39.2 & 61.6 & 64.0 & 43.2 & 67.7 & 80.7 \\
\textbf{MR-CenterNet}~$\dagger$ & Swin-L & 57.1 & 73.7 & 62.4 & 38.7 & 59.2 & 71.3 & 40.9 & 67.4 & 72.2 & 54.8 & 75.1 & 86.8\\
\hline
\end{tabular}}
\label{tab:sota}
\end{table*}

\begin{figure*}[t]
  \subfigure{ 
    \includegraphics[height=0.122\textwidth,width=0.07\textheight]{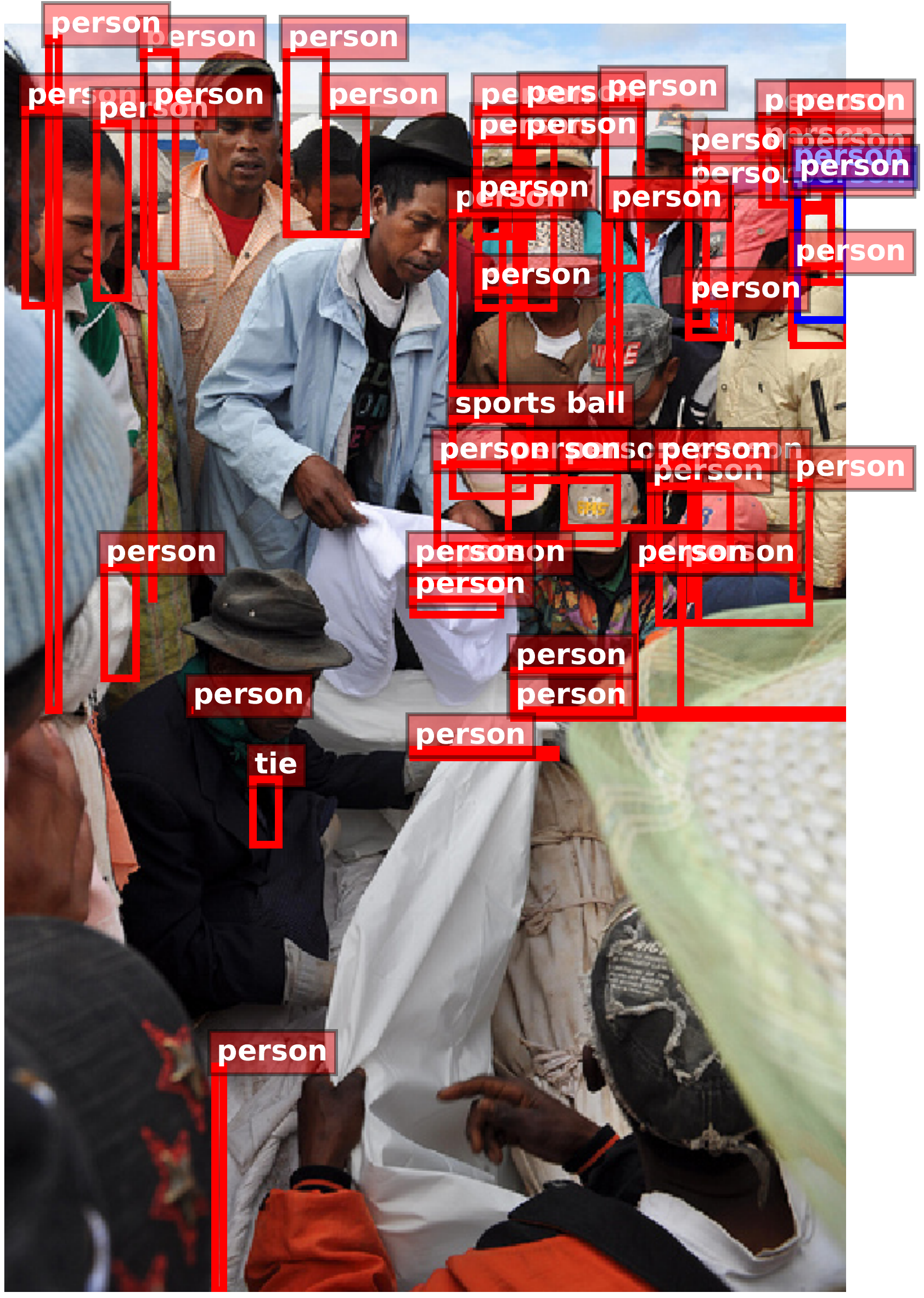}
    \label{fig:fig7_11}
  } 
  \hspace{-0.1in}
  \subfigure{ 
    \includegraphics[height=0.12\textwidth,width=0.13\textheight]{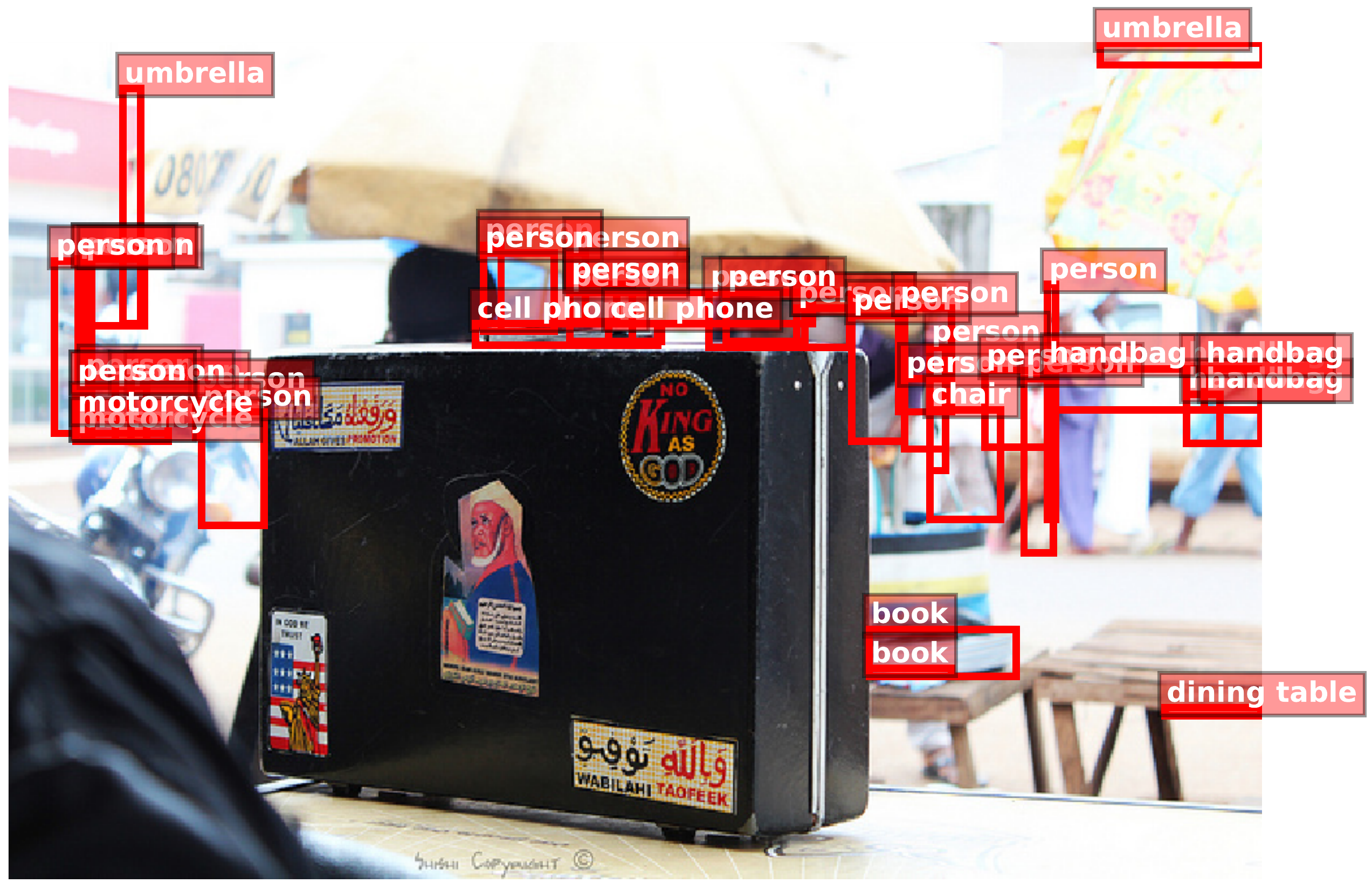}
    \label{fig:fig7_21}
  } 
  \hspace{-0.11in}
  \subfigure{  
    \includegraphics[height=0.125\textwidth,width=0.12\textheight]{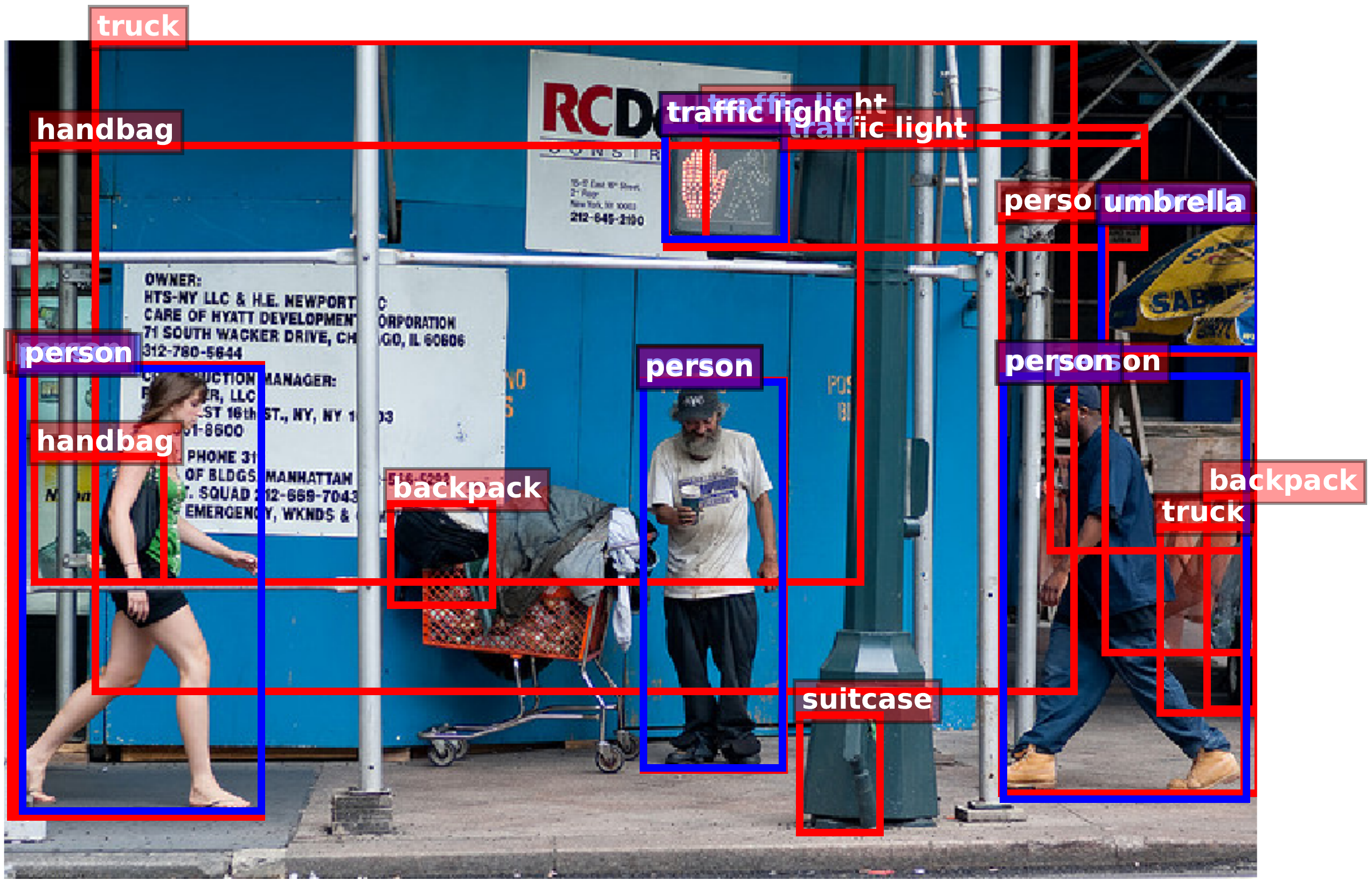}
    \label{fig:fig7_31}
  } 
  \hspace{-0.11in}
  \subfigure{ 
    \includegraphics[height=0.12\textwidth,width=0.12\textheight]{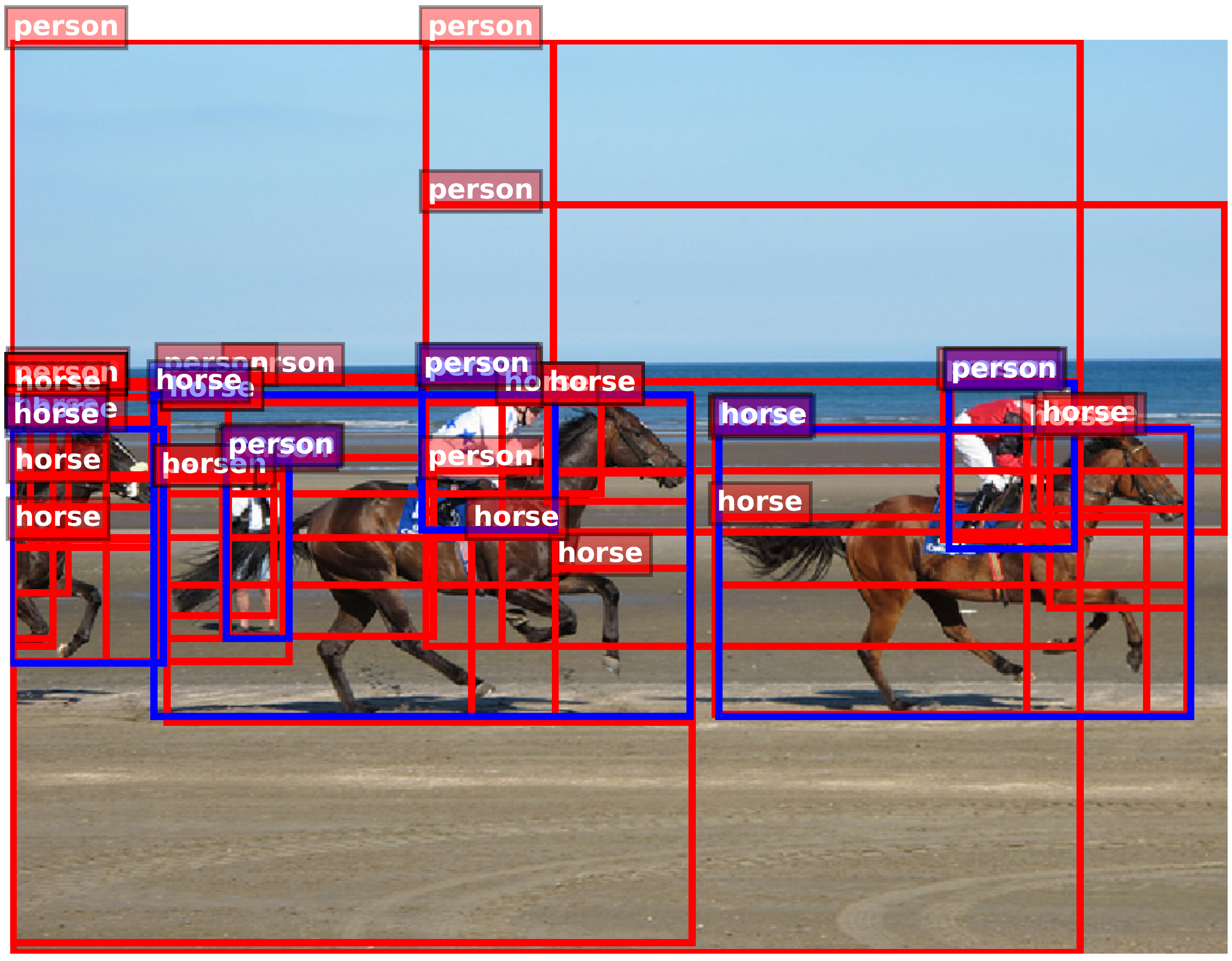}
    \label{fig:fig7_41}
  } 
  \hspace{-0.11in}
  \subfigure{ 
    \includegraphics[height=0.12\textwidth,width=0.12\textheight]{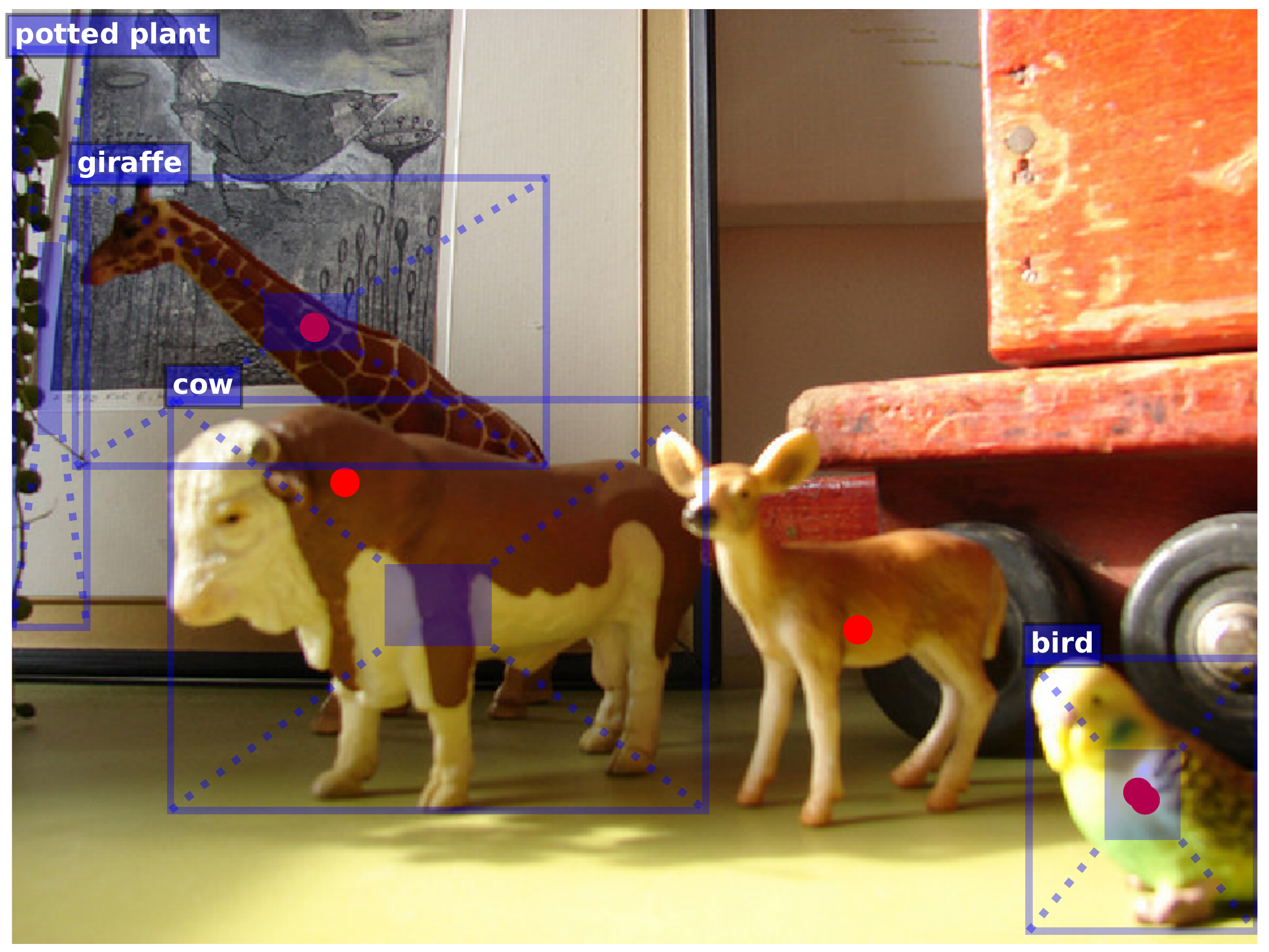}
    \label{fig:fig7_51}
  }
  \hspace{-0.11in}
  \subfigure{ 
    \includegraphics[height=0.12\textwidth,width=0.12\textheight]{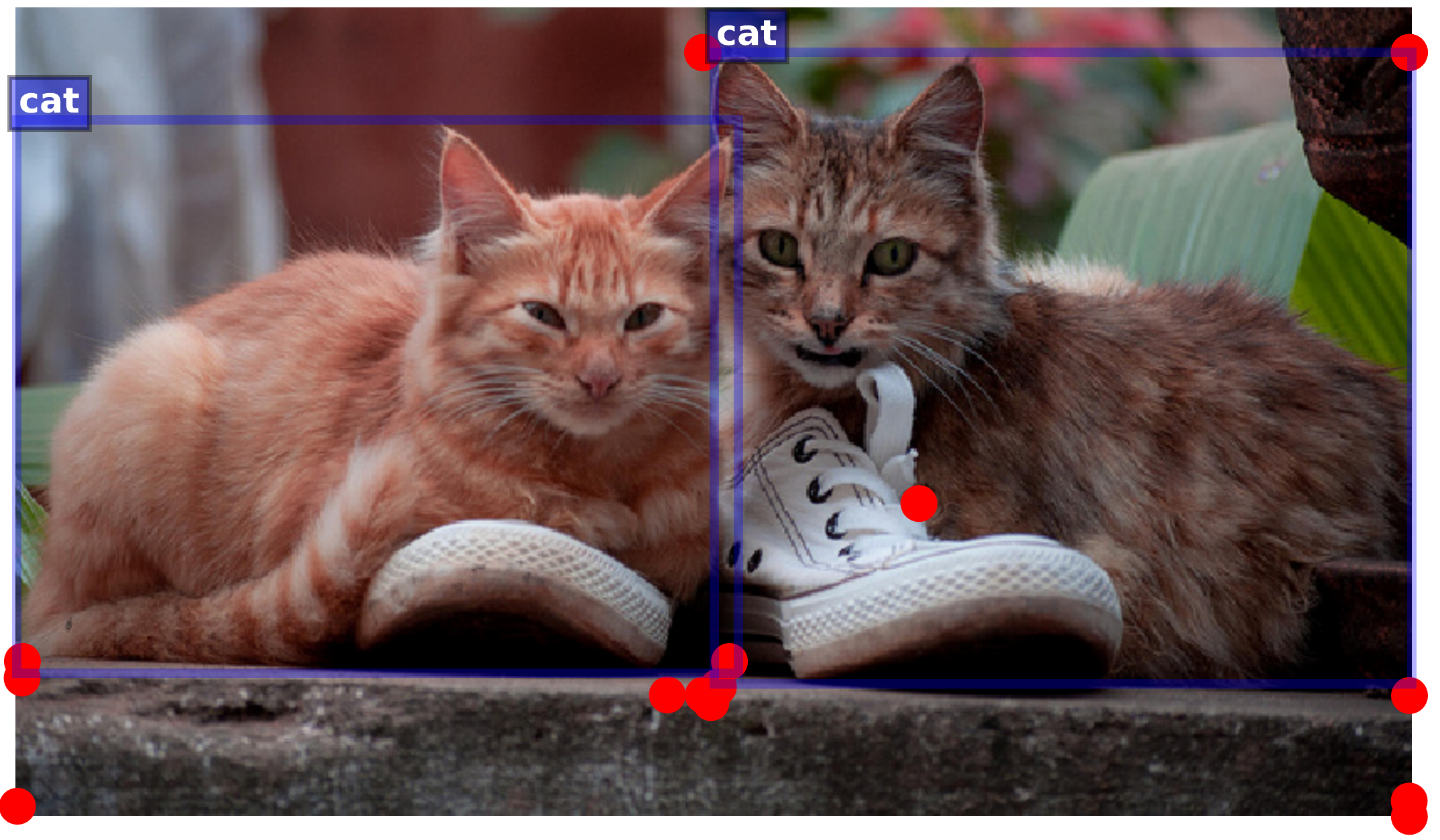}
    \label{fig:fig7_61}
  }
  \renewcommand\thesubfigure{(a)}
  \hspace{-0.1in}
  \subfigure[]{ 
    \includegraphics[height=0.122\textwidth,width=0.071\textheight]{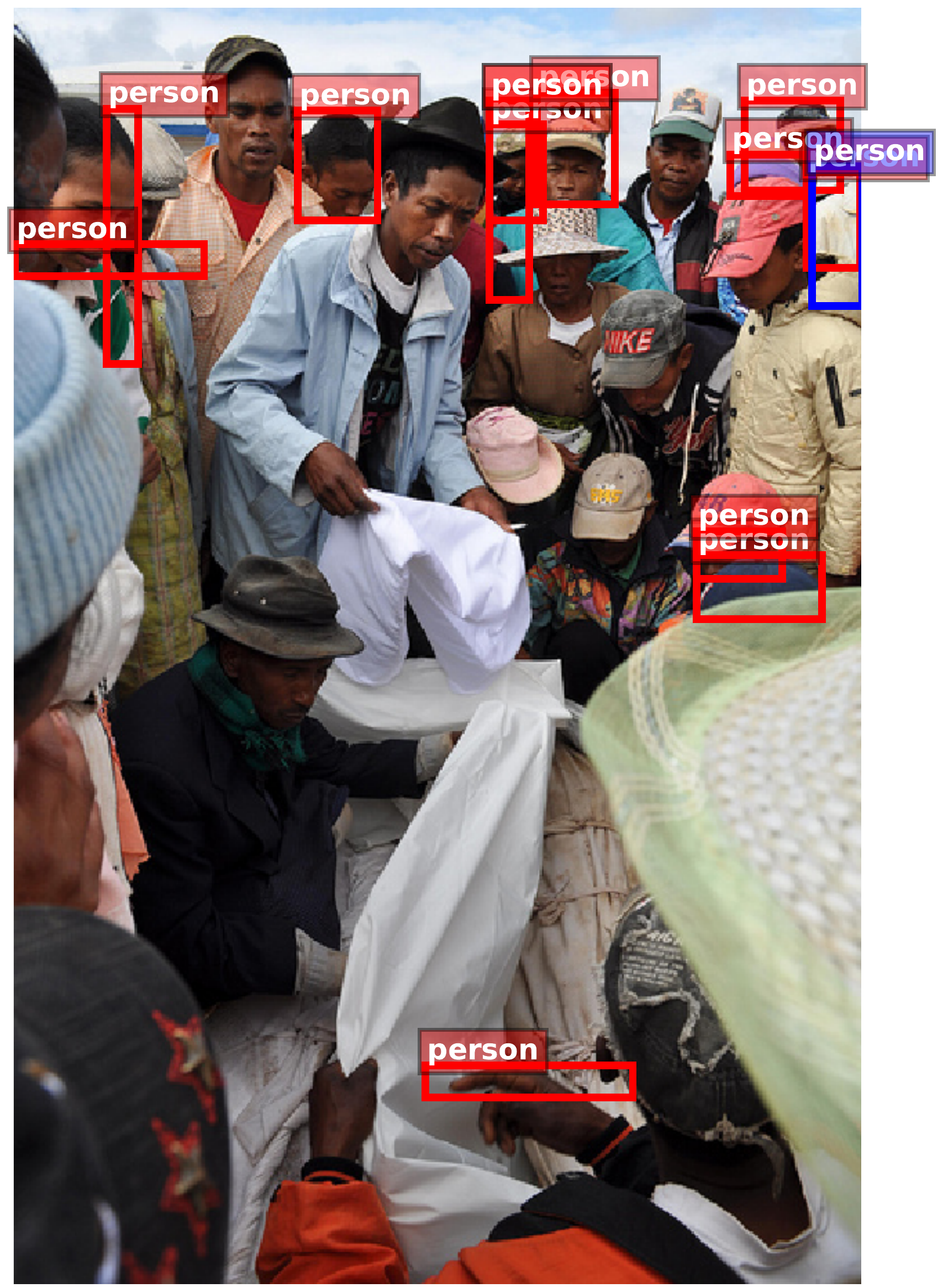}
    \label{fig:fig7_12}
  } 
  \renewcommand\thesubfigure{(b)}
  \hspace{-0.11in}
  \subfigure[]{ 
    \includegraphics[height=0.12\textwidth,width=0.122\textheight]{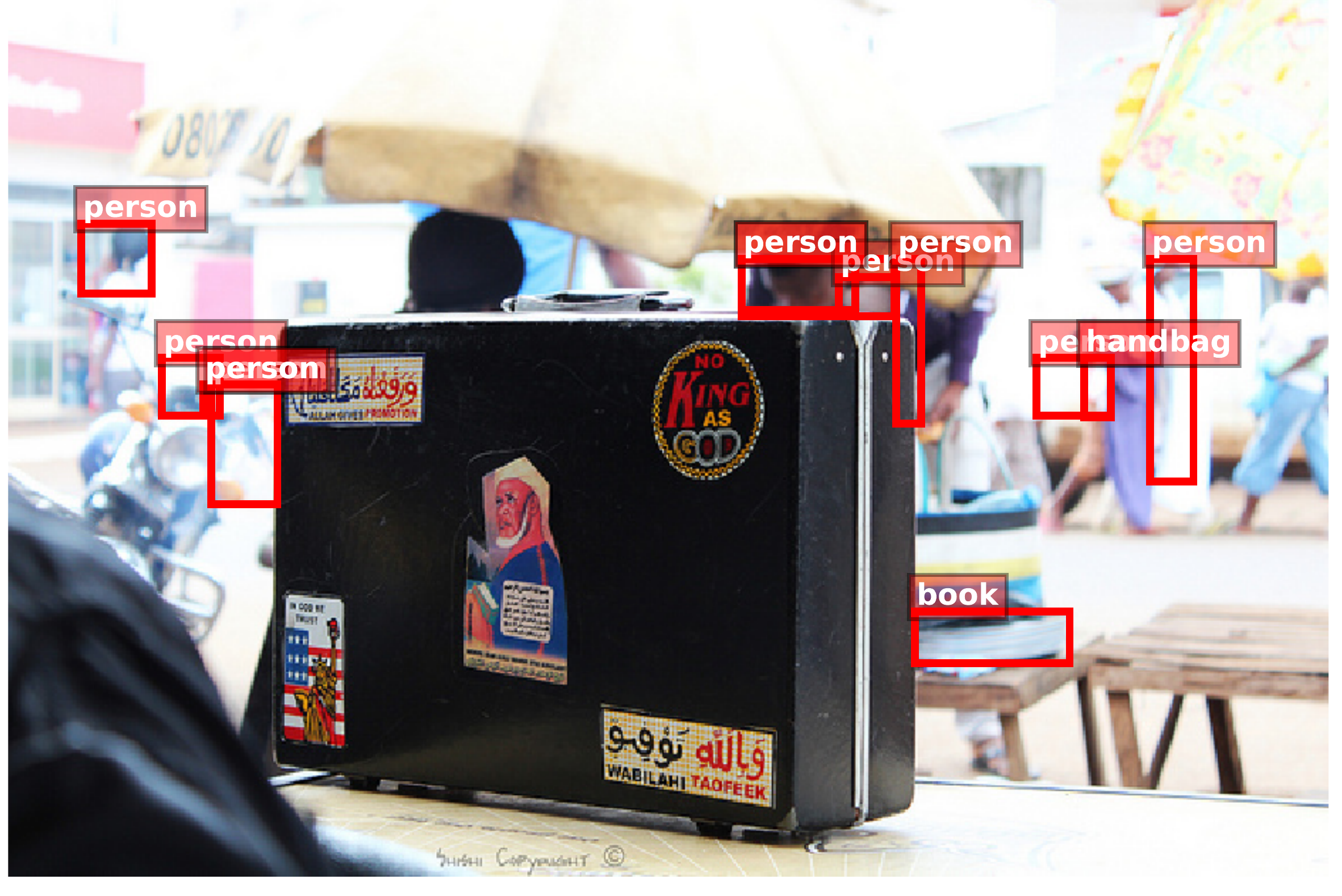}
    \label{fig:fig7_22}
  } 
  \renewcommand\thesubfigure{(c)}
  \hspace{-0.04in}
  \subfigure[]{ 
    \includegraphics[height=0.12\textwidth,width=0.12\textheight]{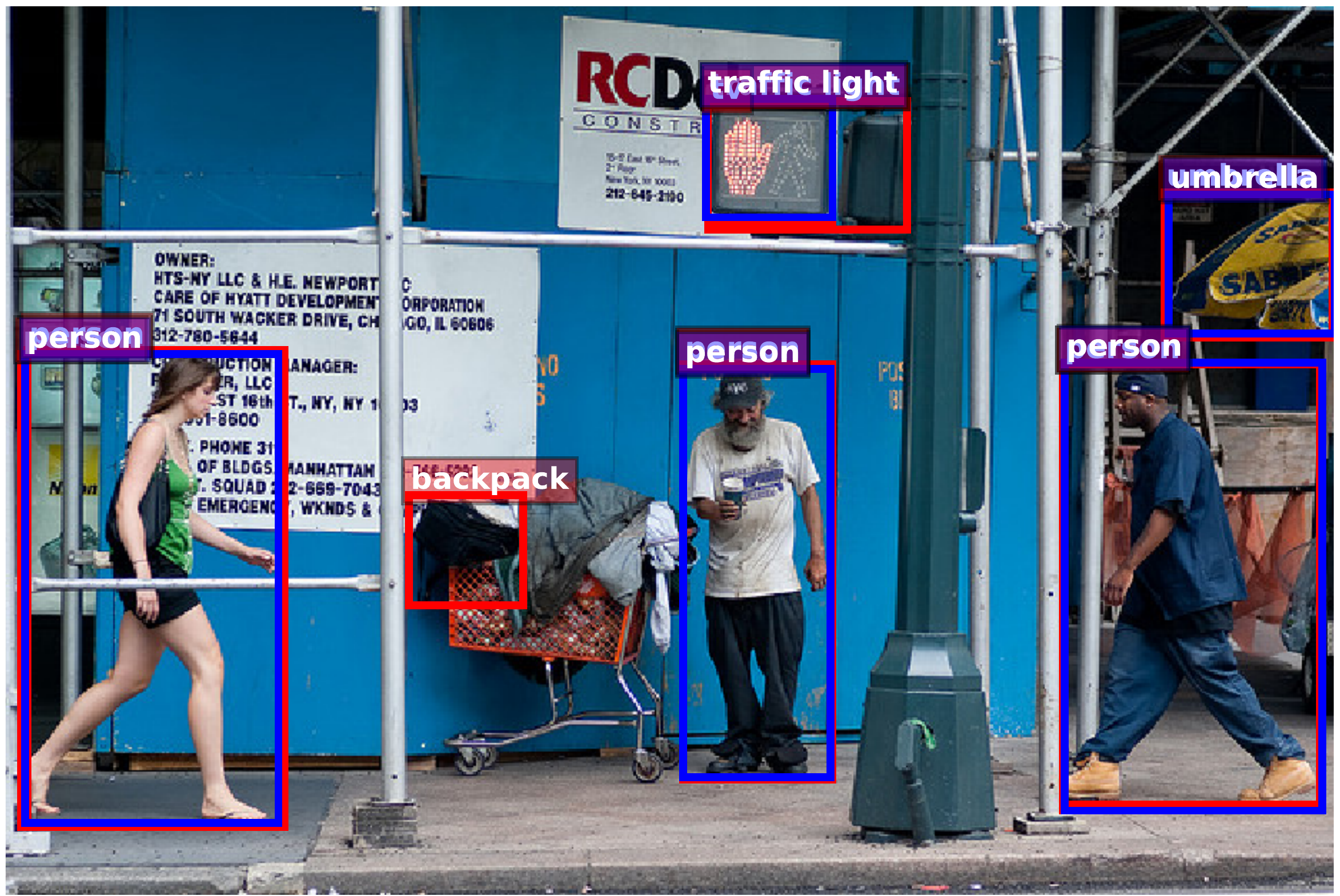}
    \label{fig:fig7_32}
  } 
  \renewcommand\thesubfigure{(d)}
  \hspace{-0.02in}
  \subfigure[]{ 
    \includegraphics[height=0.12\textwidth,width=0.12\textheight]{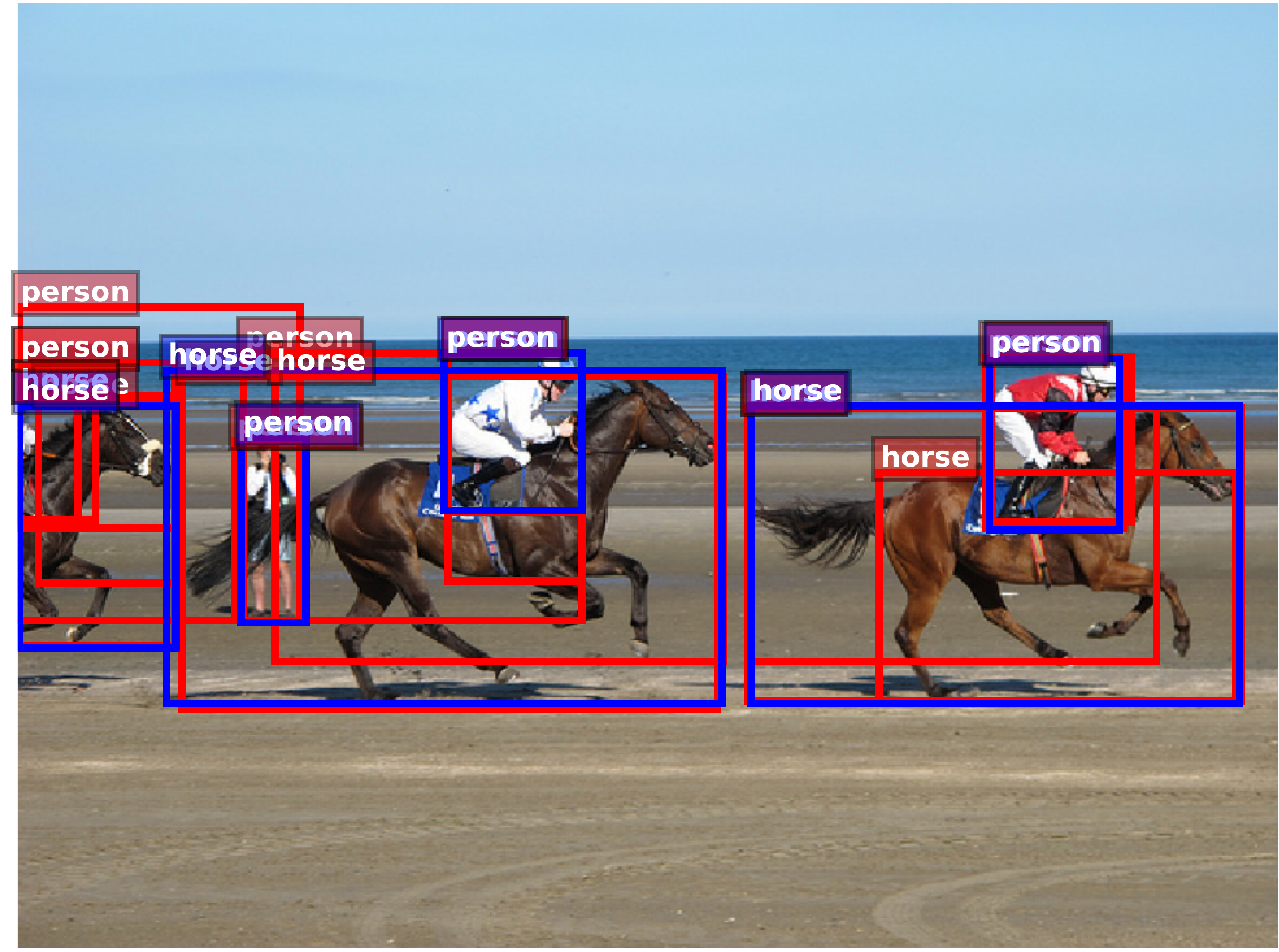}
    \label{fig:fig7_42}
  }
  \renewcommand\thesubfigure{(e)}
  \hspace{-0.1in}
  \subfigure[]{ 
    \includegraphics[height=0.12\textwidth,width=0.12\textheight]{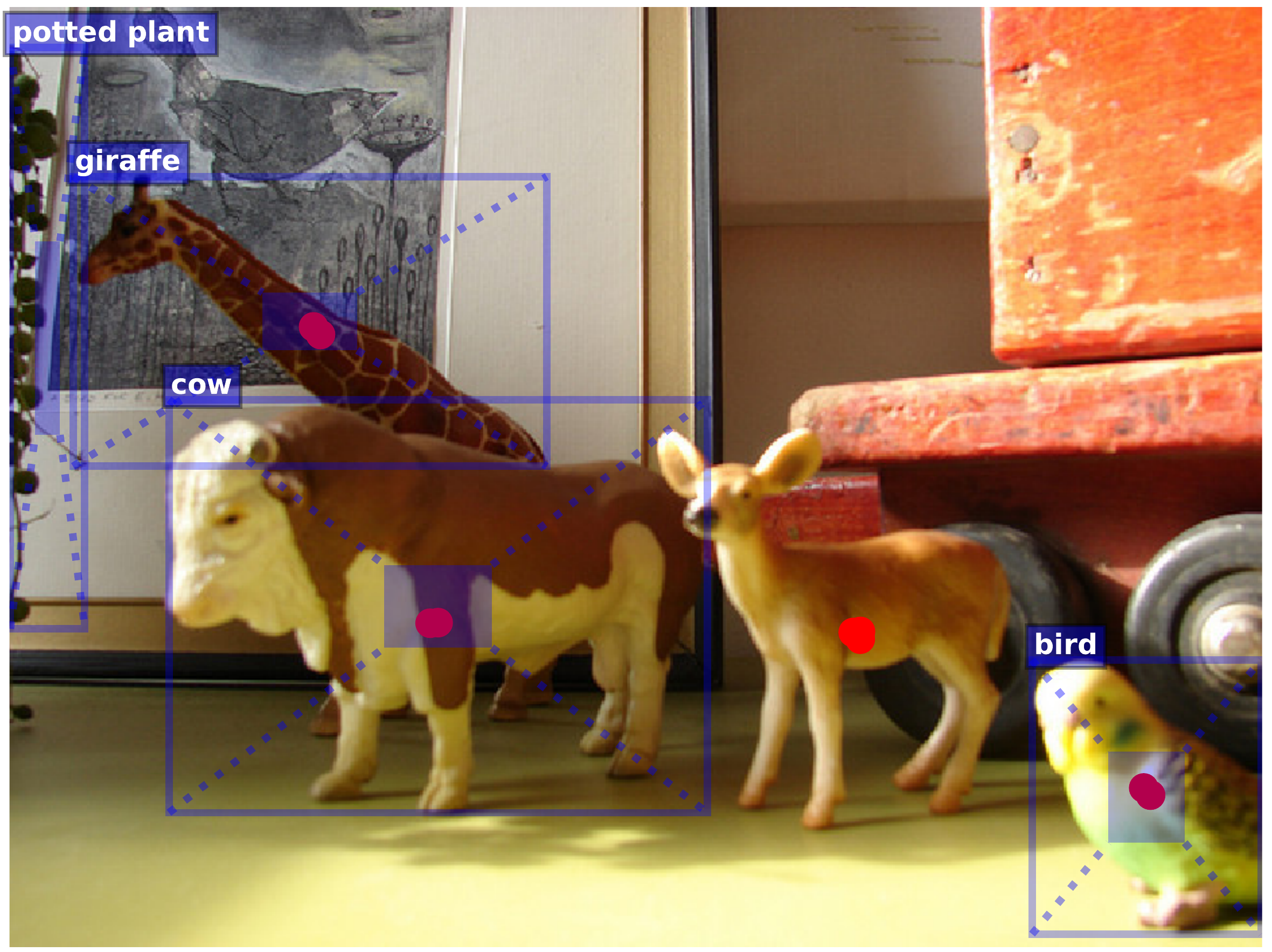}
    \label{fig:fig7_52}
  } 
  \renewcommand\thesubfigure{(f)}
  \hspace{-0.1in}
  \subfigure[]{ 
    \includegraphics[height=0.12\textwidth,width=0.12\textheight]{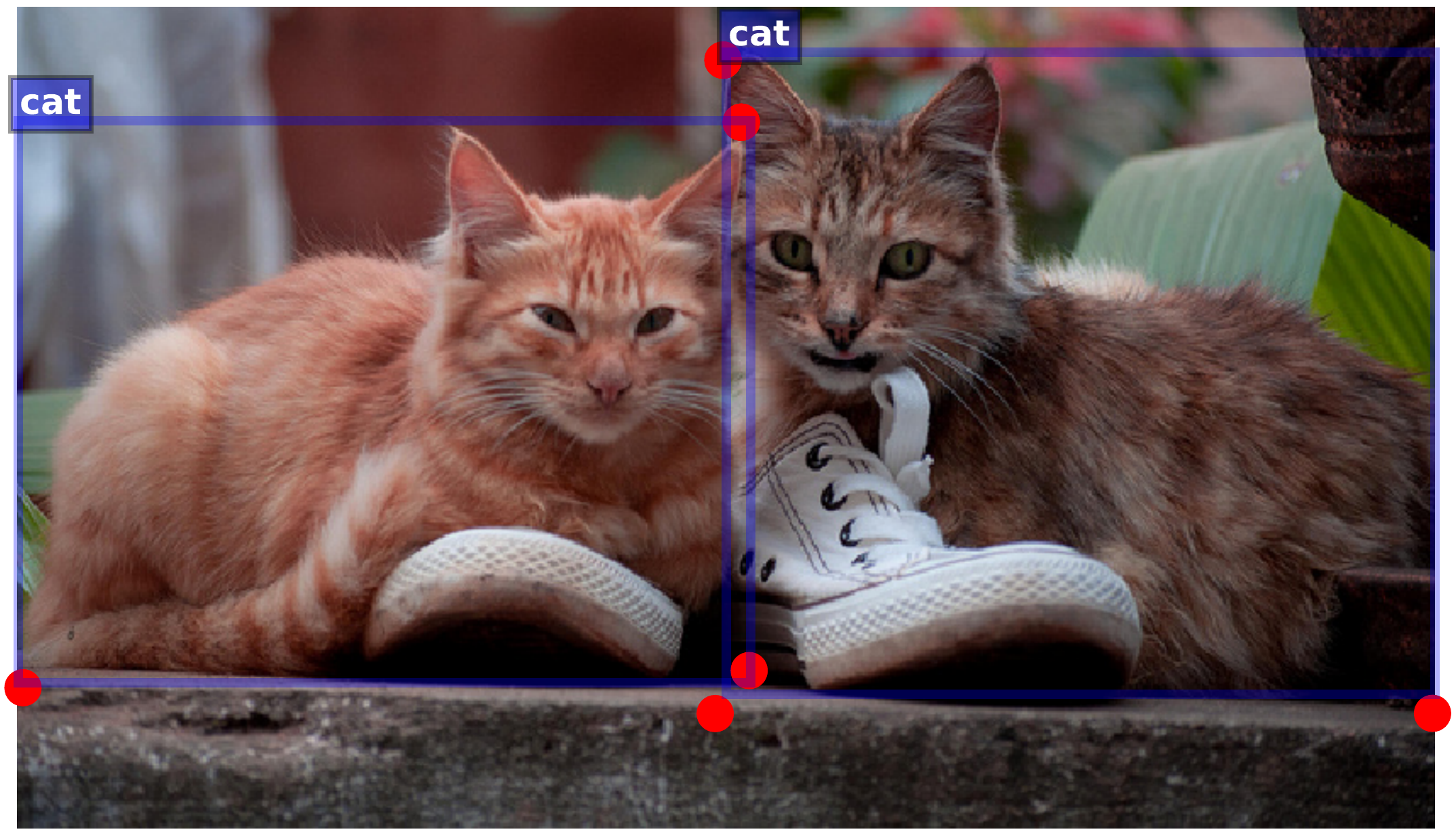}
    \label{fig:fig7_62}
  }
  \caption{(a) and (b) show that the small incorrect bounding boxes are significantly reduced by the modeling center information. (c) and (d) show that the center information works to reduce the medium and large incorrect bounding boxes. (e) shows the results of the detection of the center keypoints without/with center pooling. (f) shows the results of the detection of corners with corner pooling and cascade corner pooling. The blue boxes above denote the ground-truth. The red boxes and dots denote the predicted bounding boxes and keypoints, respectively.} 
  \label{fig:qualitative}
\end{figure*}

\begin{figure*}[t]
  \centering 
  \subfigure{ 
    \includegraphics[height=0.12\textwidth,width=0.13\textheight]{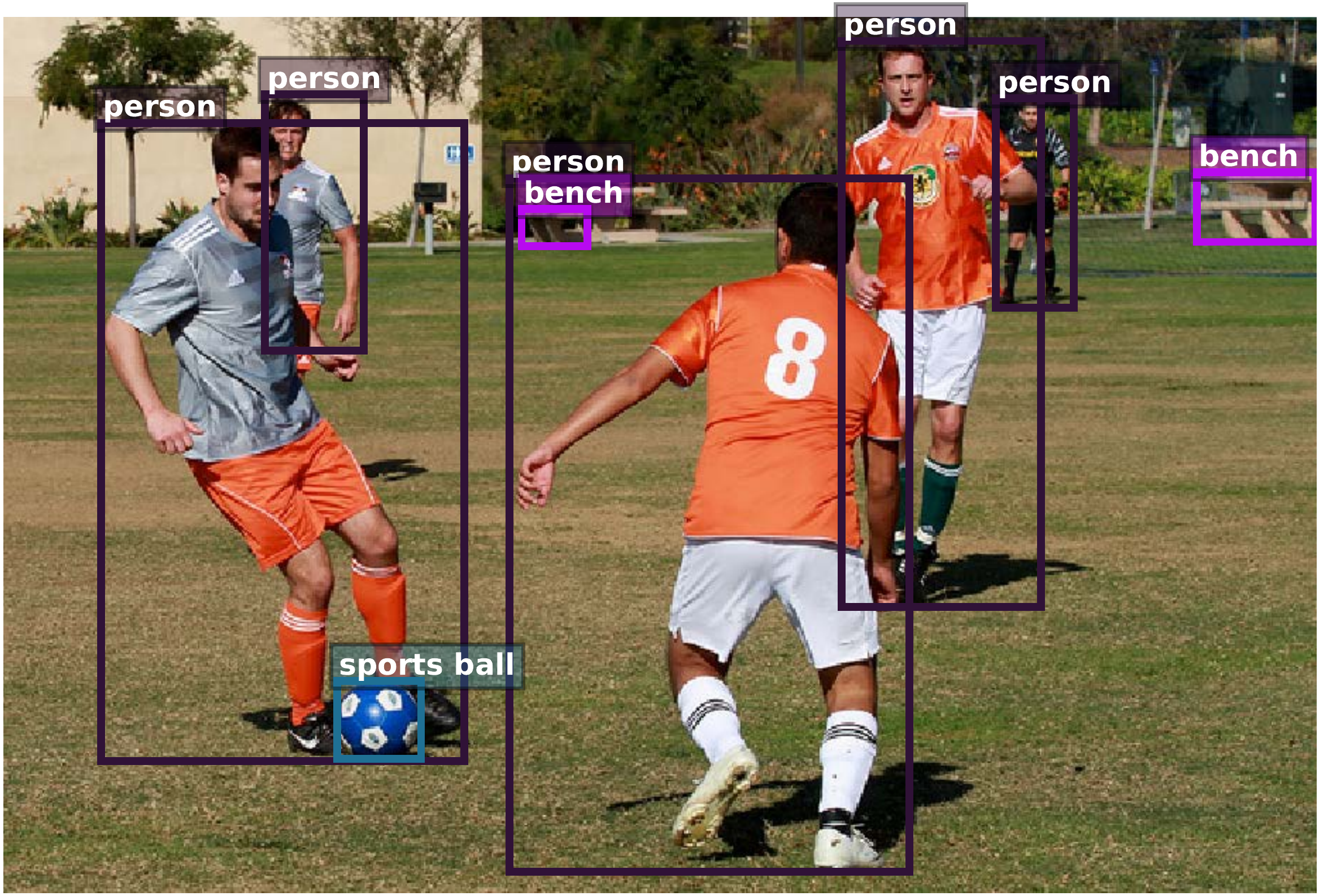}
    \label{fig8_1}
  }
  \hspace{-0.05in}
  \subfigure{ 
    \includegraphics[height=0.12\textwidth,width=0.08\textheight]{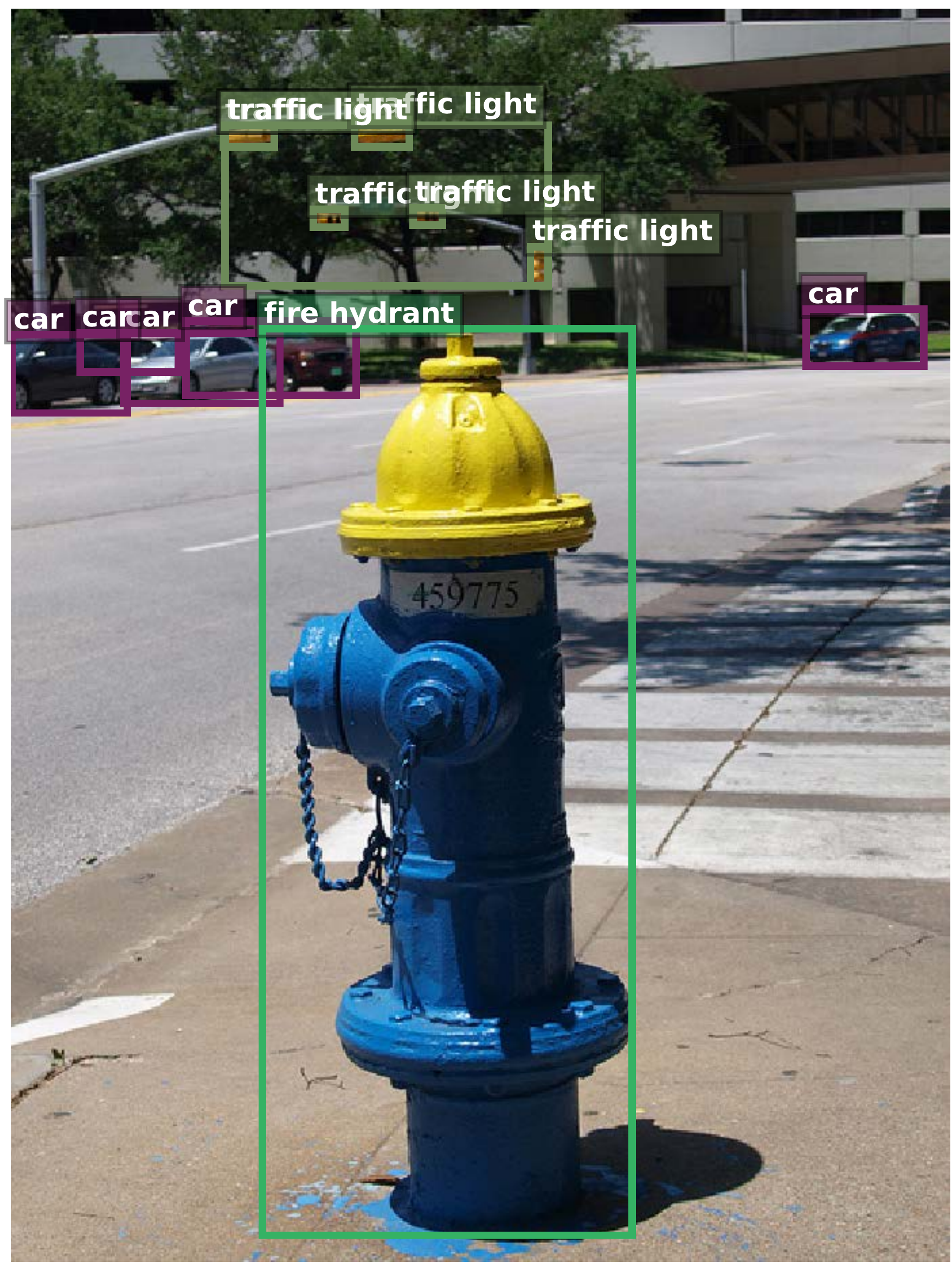}
    \label{fig8_2}
  }
  \hspace{-0.05in}
  \subfigure{ 
    \includegraphics[height=0.12\textwidth,width=0.13\textheight]{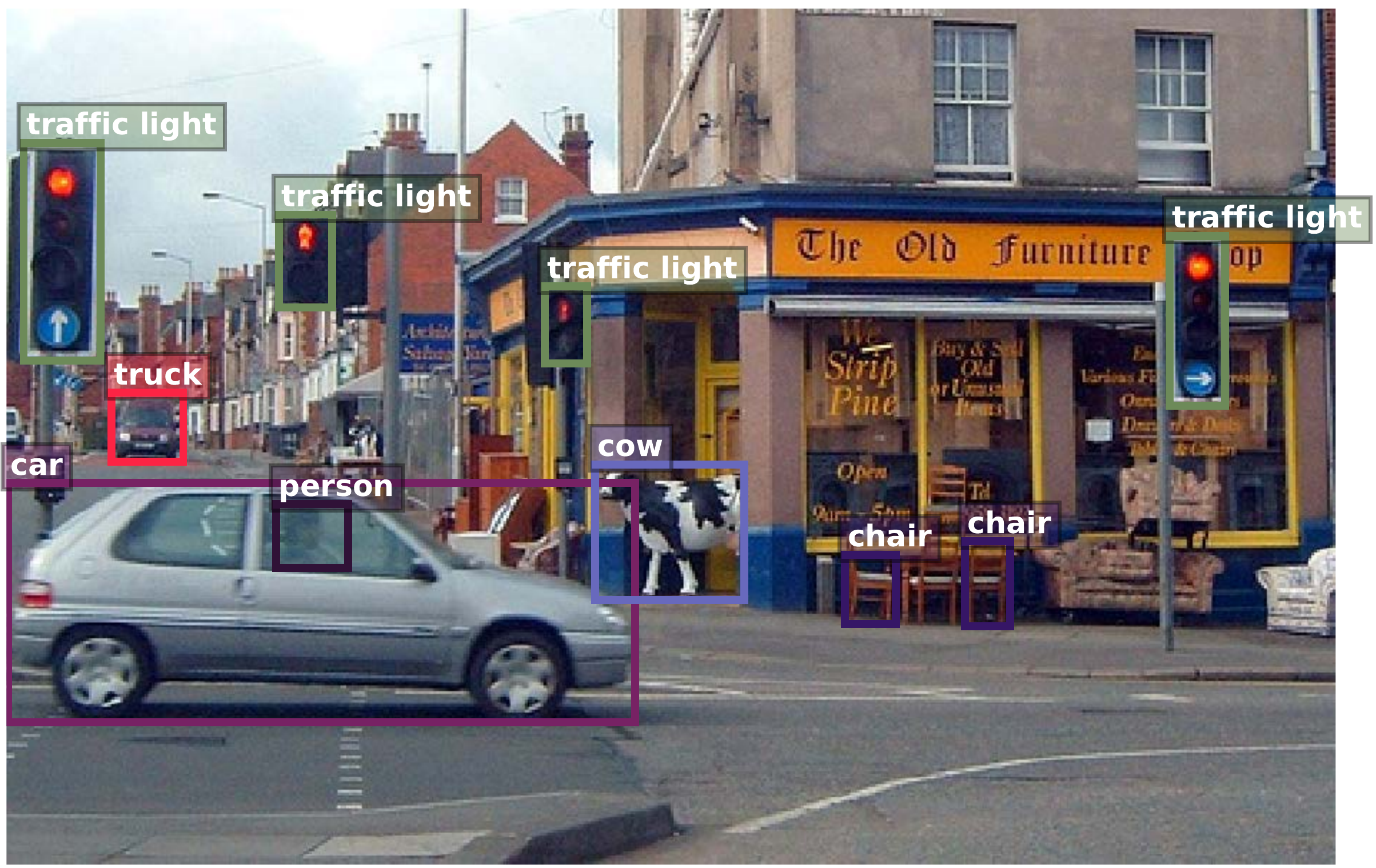}
    \label{fig8_3}
  } 
  \hspace{-0.05in}
  \subfigure{ 
    \includegraphics[height=0.12\textwidth,width=0.12\textheight]{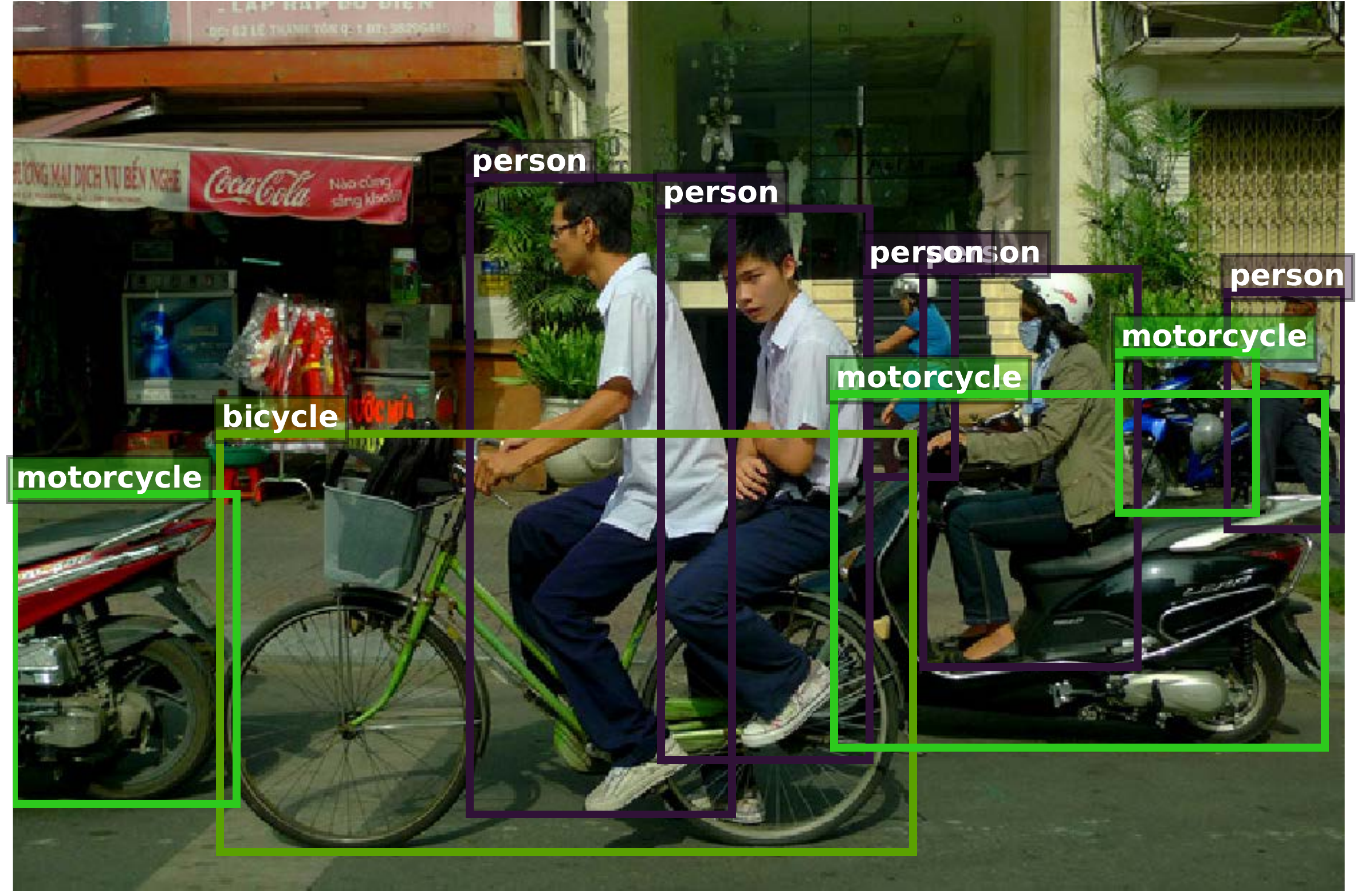}
    \label{fig8_4}
  } 
  \hspace{-0.05in}
  \subfigure{ 
    \includegraphics[height=0.12\textwidth,width=0.08\textheight]{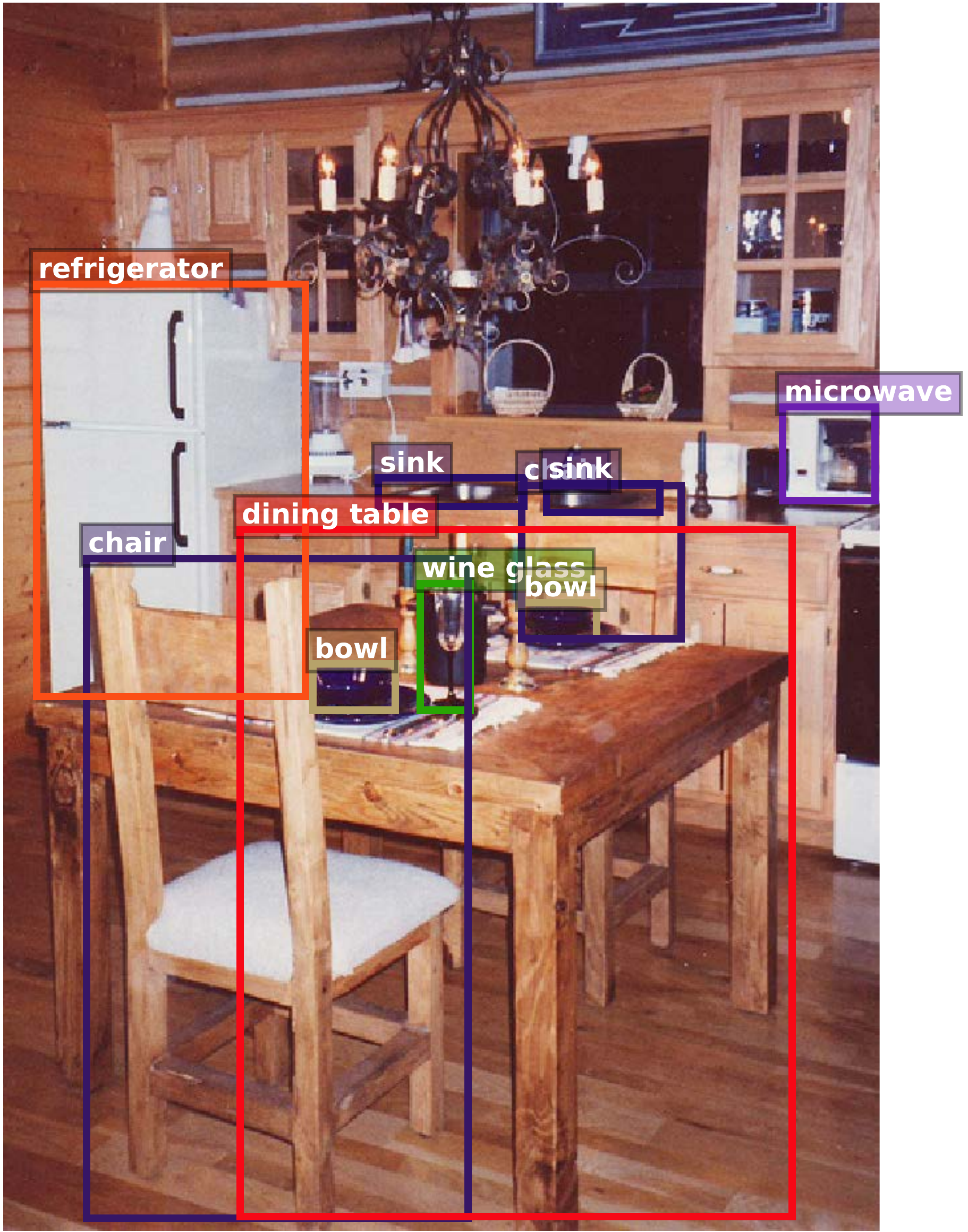}
    \label{fig8_5}
  } 
  \hspace{-0.05in}
  \subfigure{ 
    \includegraphics[height=0.125\textwidth,width=0.12\textheight]{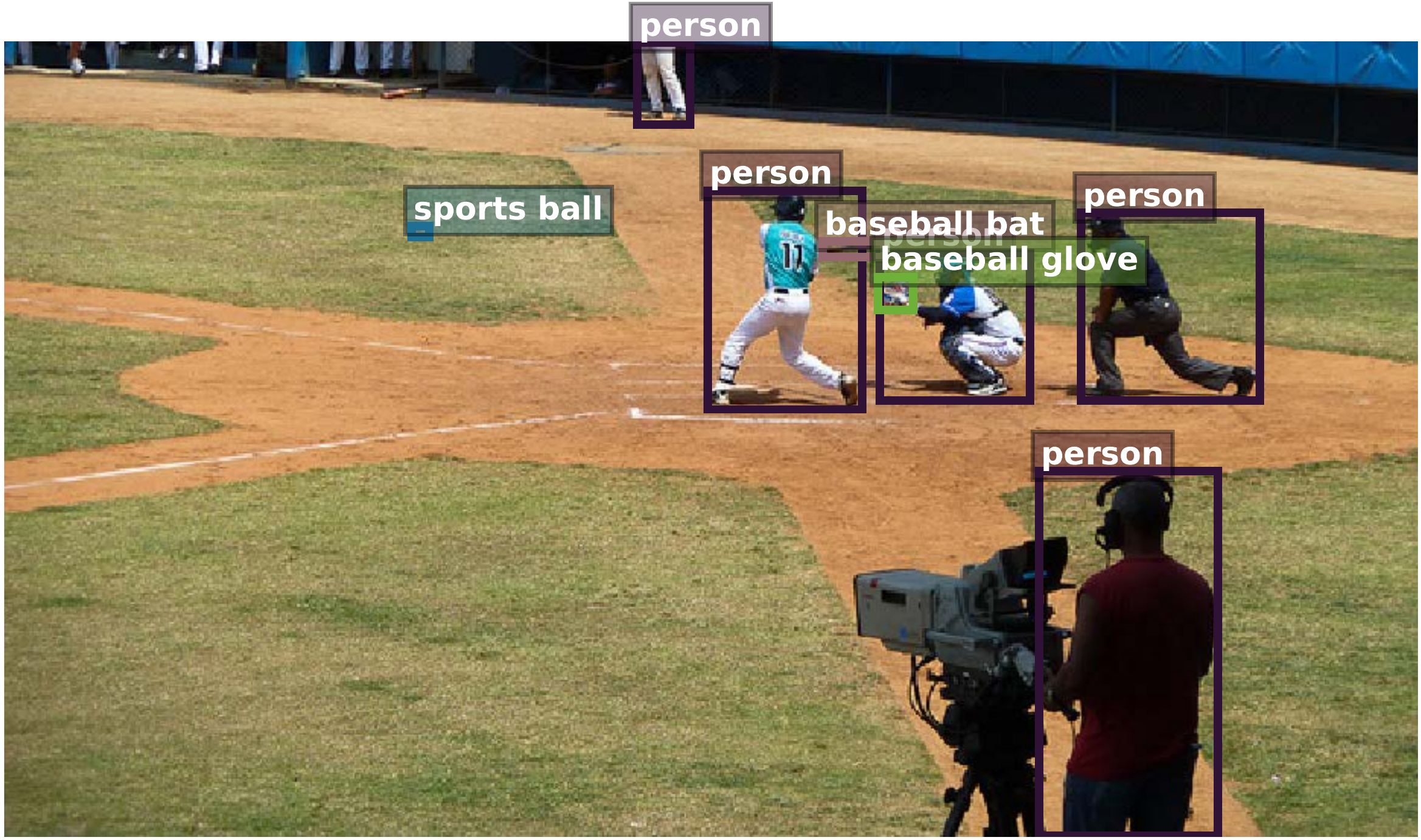}
    \label{fig8_6}
  }
  \subfigure{ 
    \includegraphics[height=0.12\textwidth,width=0.129\textheight]{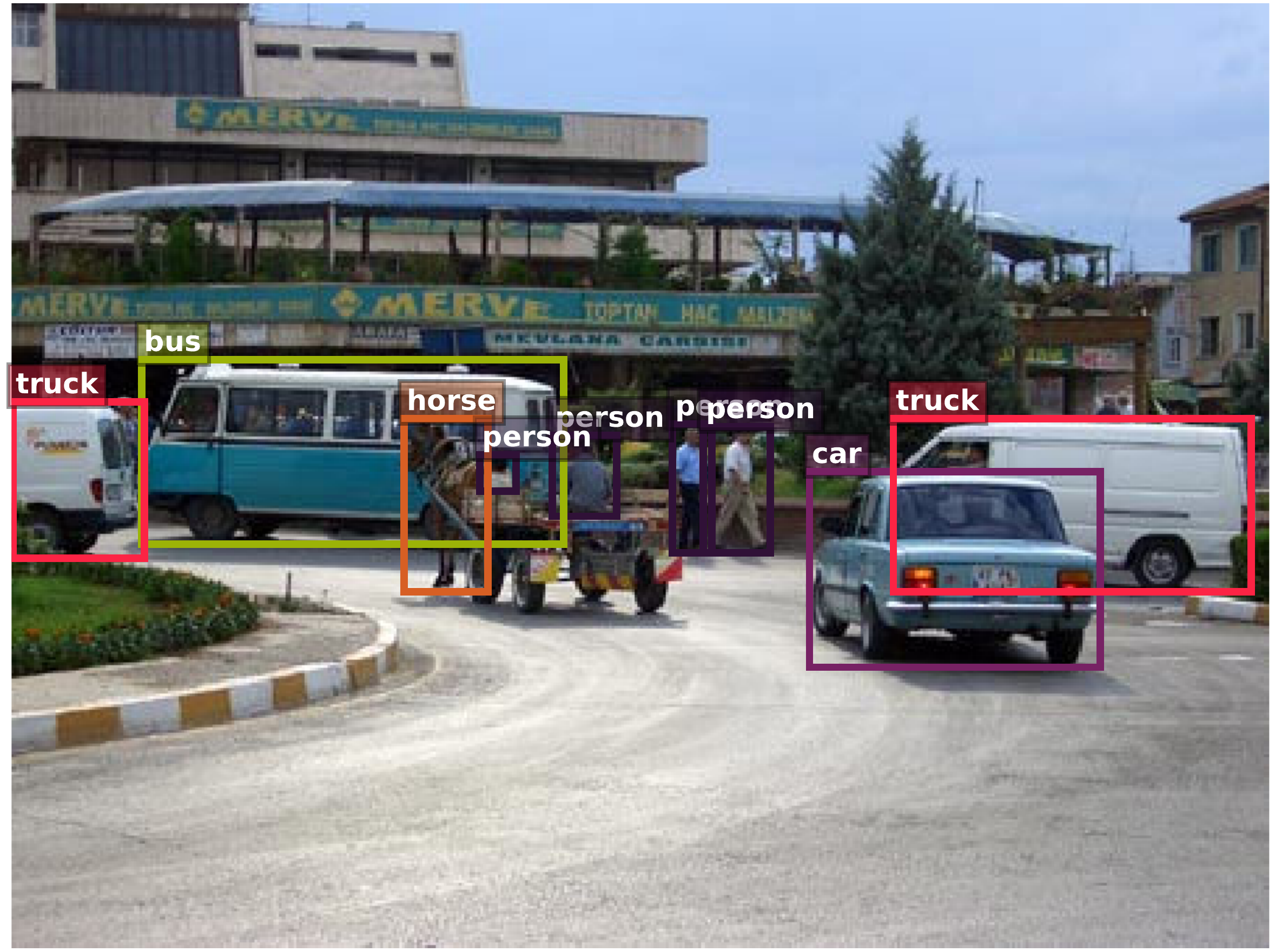}
    \label{fig8_137}
  }
  \hspace{-0.05in}
  \subfigure{ 
    \includegraphics[height=0.123\textwidth,width=0.085\textheight]{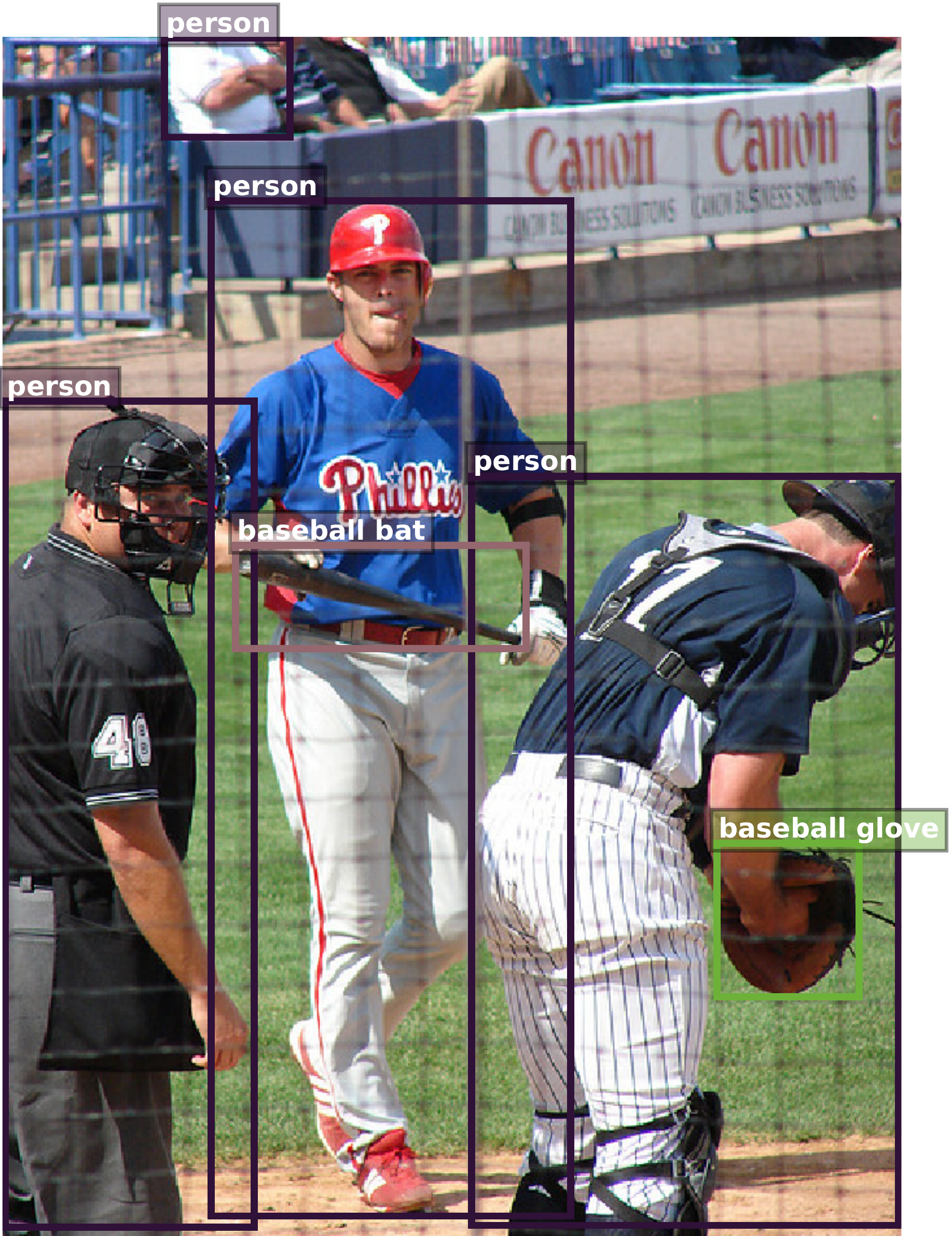}
    \label{fig8_231}
  }
  \hspace{-0.1in}
  \subfigure{ 
    \includegraphics[height=0.12\textwidth,width=0.134\textheight]{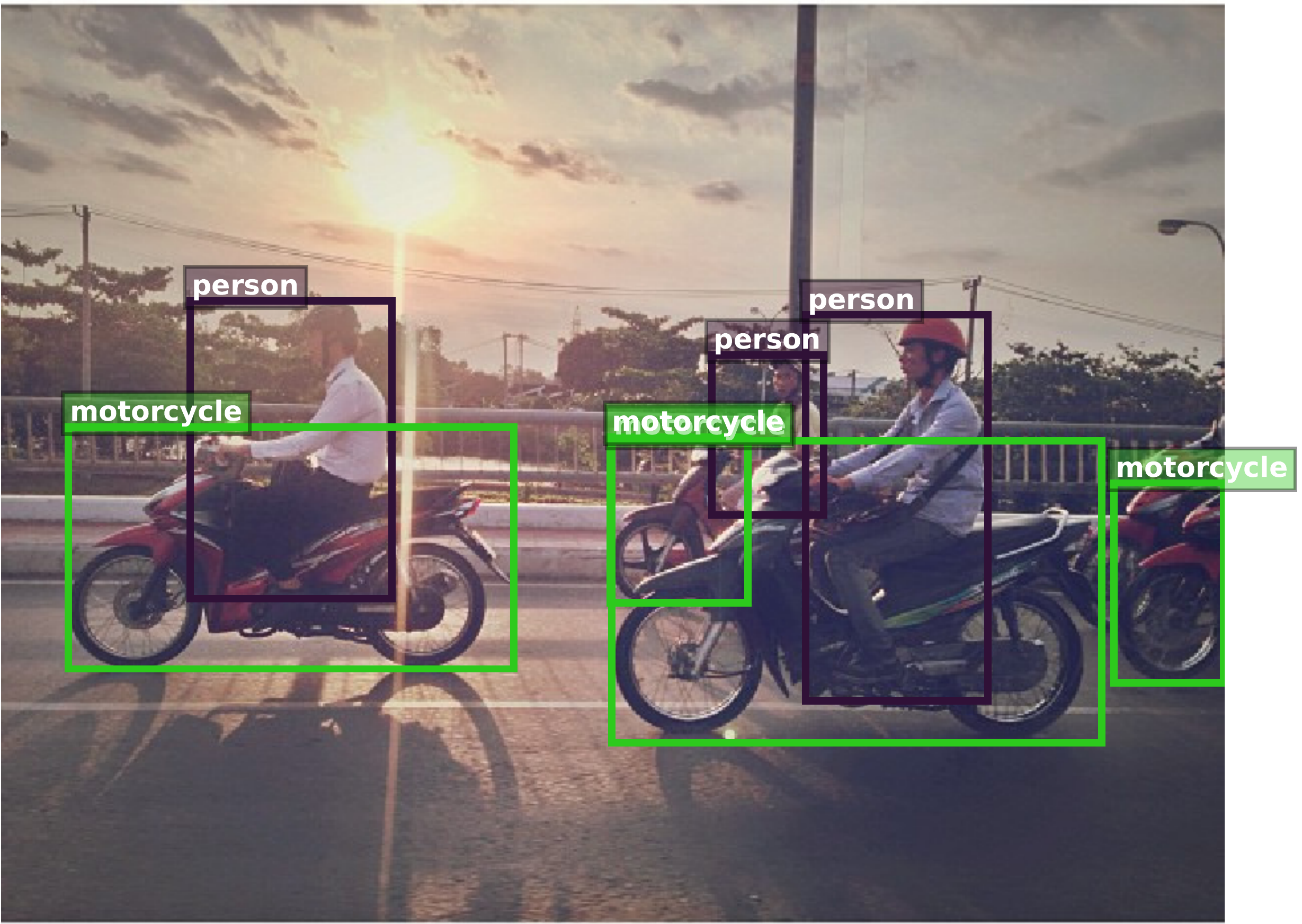}
    \label{fig8_73}
  }
  \hspace{-0.1in}
  \subfigure{ 
    \includegraphics[height=0.125\textwidth,width=0.121\textheight]{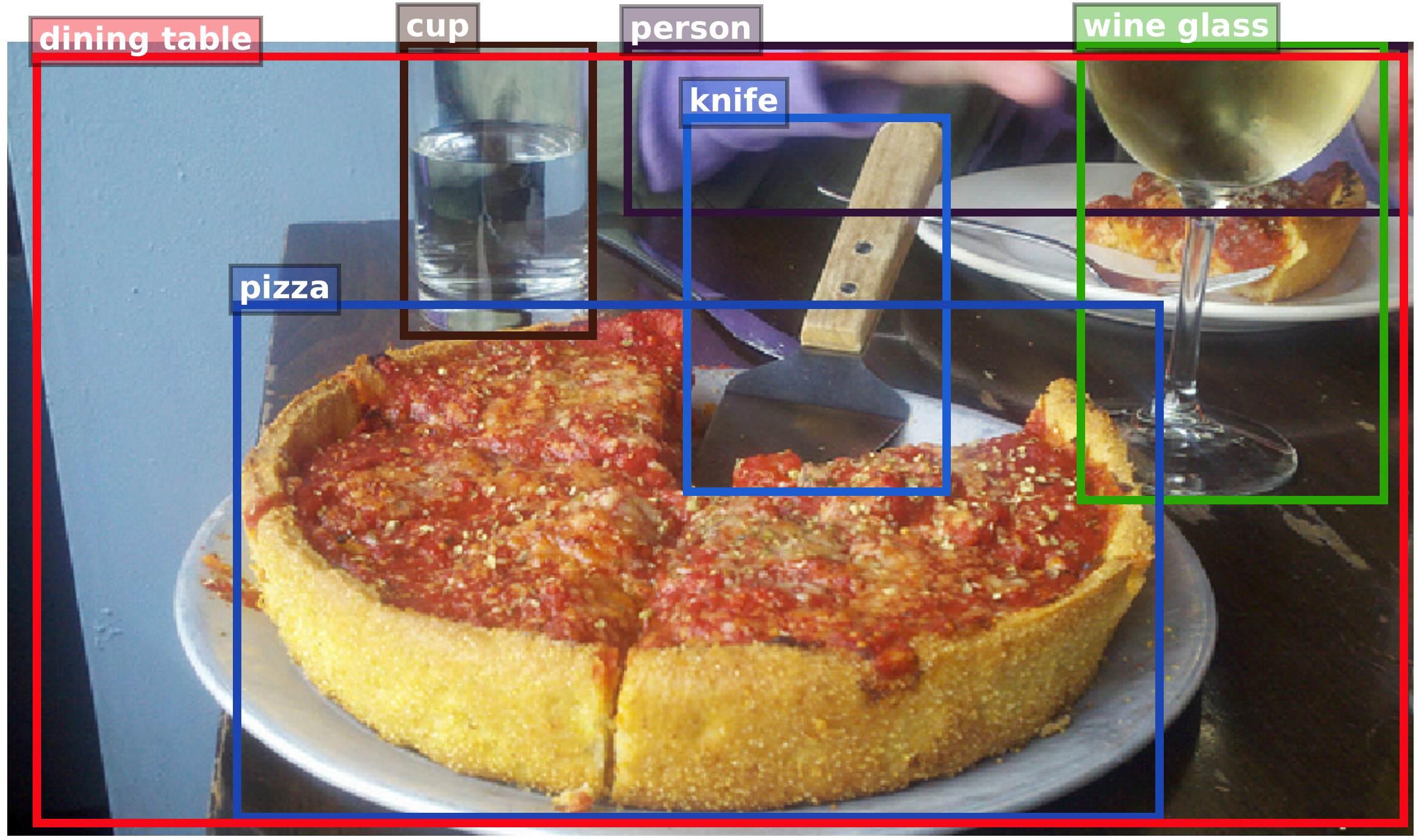}
    \label{fig8_220}
  }
  \hspace{-0.05in}
  \subfigure{ 
    \includegraphics[height=0.12\textwidth,width=0.075\textheight]{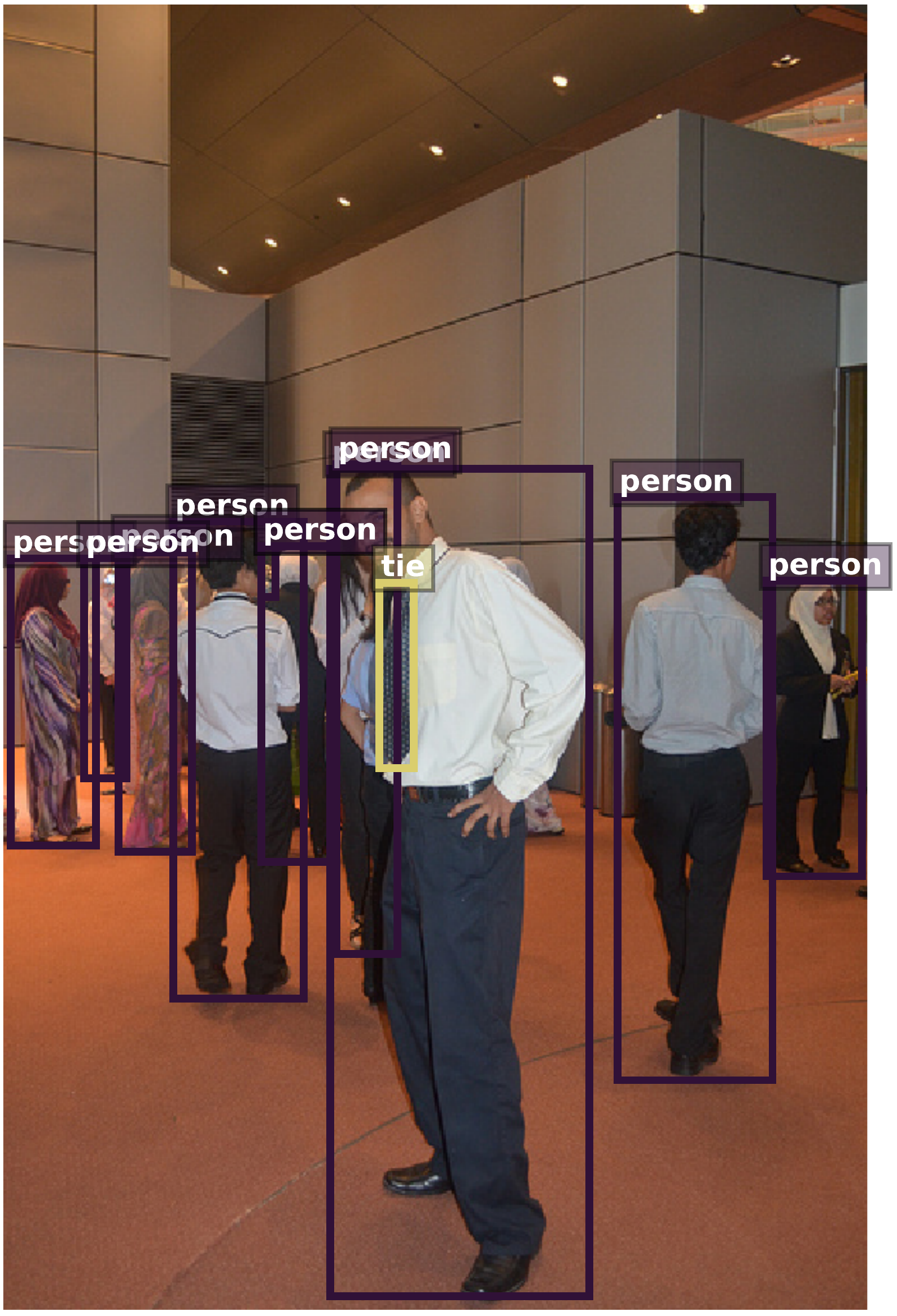}
    \label{fig8_173}
  }
  \subfigure{ 
    \includegraphics[height=0.123\textwidth,width=0.12\textheight]{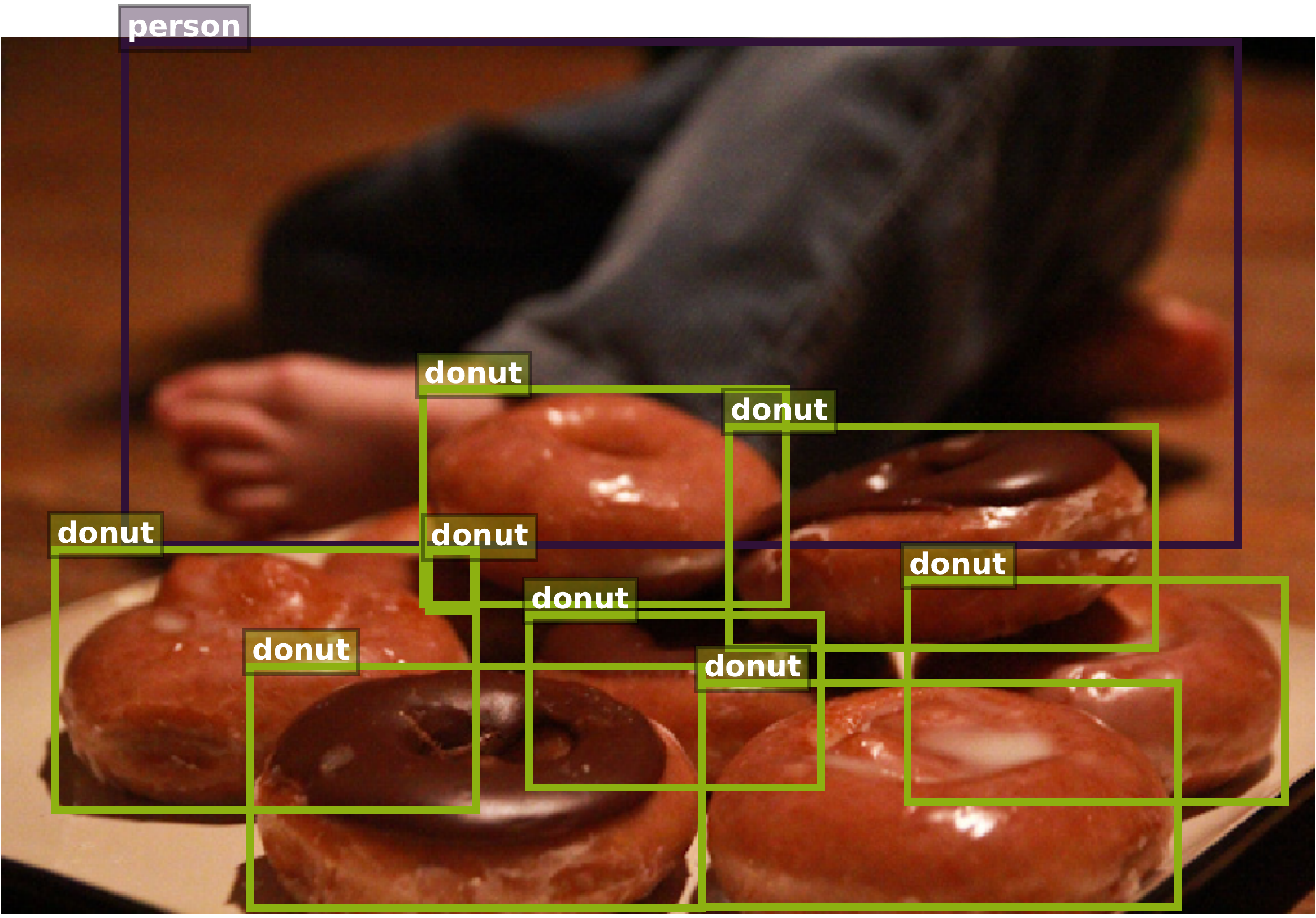}
    \label{fig8_254}
  }
  \caption{Some qualitative detection results on the MS-COCO validation dataset. Only detections with scores higher than $0.5$ are shown.} 
  \label{qualitative_detection}
\end{figure*}

Table~\ref{tab:sota} shows a comparison with state-of-the-art detectors on the MS-COCO test-dev set. Compared with the baseline CornerNet~\cite{law2018cornernet}, the proposed CenterNet achieves a remarkable improvement. For example, SR-CenterNet (Hourglass-52) reports a single-scale testing AP of $41.6\%$, an improvement of $3.8\%$ over $37.8\%$, and a multi-scale testing AP of $43.5\%$, an improvement of $4.1\%$ over $39.4\%$, achieved by CornerNet under the same setting. When using the deeper backbone (\ie, Hourglass-104), the AP improvement over CornerNet are $4.4\%$ (from $40.5\%$ to $44.9\%$) and $4.9\%$ (from $42.1\%$ to $47.0\%$) under the single-scale and multi-scale testing, respectively. We also report the detection results of MR-CenterNet, which obtain significant improvements again on the basis of SR-CenterNet, \eg, with the backbone of Res2Net-101, MR-CenterNet reports APs of $51.5\%$ for single-scale testing and $53.7\%$ for multi-scale testing, respectively. The current Transformer based top-down detectors~\cite{liu2021swin} push the accuracy to a new level, so we explore the application of the Transformer based backbone to the bottom-up approaches. After equipped with the Transformer based backbone, CenterNet reports APs of $53.2\%$ for single-scale testing and $57.1\%$ for multi-scale testing, respectively, surpassing all other bottom-up approaches to the best of our knowledge.

The biggest improvement comes from small objects. For instance, SR-CenterNet (Hourglass-52) improves the AP for small objects by $5.5\%$ (single-scale) and by $6.4\%$ (multi-scale). For the backbone Hourglass-104, the improvements are $6.2\%$ (single-scale) and $8.1\%$ (multi-scale), respectively. This benefit stems from the center information modeled by the center keypoints: the smaller the scale of an incorrect bounding box, the lower the probability that a center keypoint can be detected in the central region. Figure~\ref{fig:fig7_12} and Figure~\ref{fig:fig7_22} show qualitative comparisons that demonstrate the effectiveness of CenterNet in reducing small incorrect bounding boxes. 

CenterNet also leads to a large improvement in reducing medium and large incorrect bounding boxes. As Table~\ref{tab:sota} shows, SR-CenterNet (Hourglass-104) improves the single-scale testing AP by $4.7\%$ (from $42.7\%$ to $47.4\%$) and $3.5\%$ (from $53.9\%$ to $57.4\%$) for medium and large bounding boxes, respectively. Figure~\ref{fig:fig7_32} and Figure~\ref{fig:fig7_42} show qualitative comparisons of the reduction of medium and large incorrect bounding boxes. Notably, the AR is also significantly improved compared with baselines, with the best performance achieved with multi-scale testing. This is because our approach removes many incorrect bounding boxes, which is equivalent to improving the confidence of those bounding boxes with accurate locations but relatively low scores.

The performance of CenterNet is also competitive with those of top-down approaches,~\eg,~the single-scale testing AP of SR-CenterNet (Hourglass-52) is comparable to the top-down approach RefineDet~\cite{zhang2018single} ($41.6\%$~\emph{vs.}~$41.8\%$) and that of MR-CenterNet (Res2Net-101) is comparable to GFLV2~\cite{li2021generalized} ($53.7\%$~\emph{vs.}~$53.3\%$). The multi-scale testing AP of $57.1\%$ achieved by MR-CenterNet (Swin-L) closely matches the state-of-the-art AP of $58.7\%$ achieved by the top-down approach Swin Transformer~\cite{liu2021swin}. We present qualitative detection results in Figure~\ref{qualitative_detection}.

\begin{figure}[!tb]
  \centering 
  \includegraphics[width=0.45\textwidth]{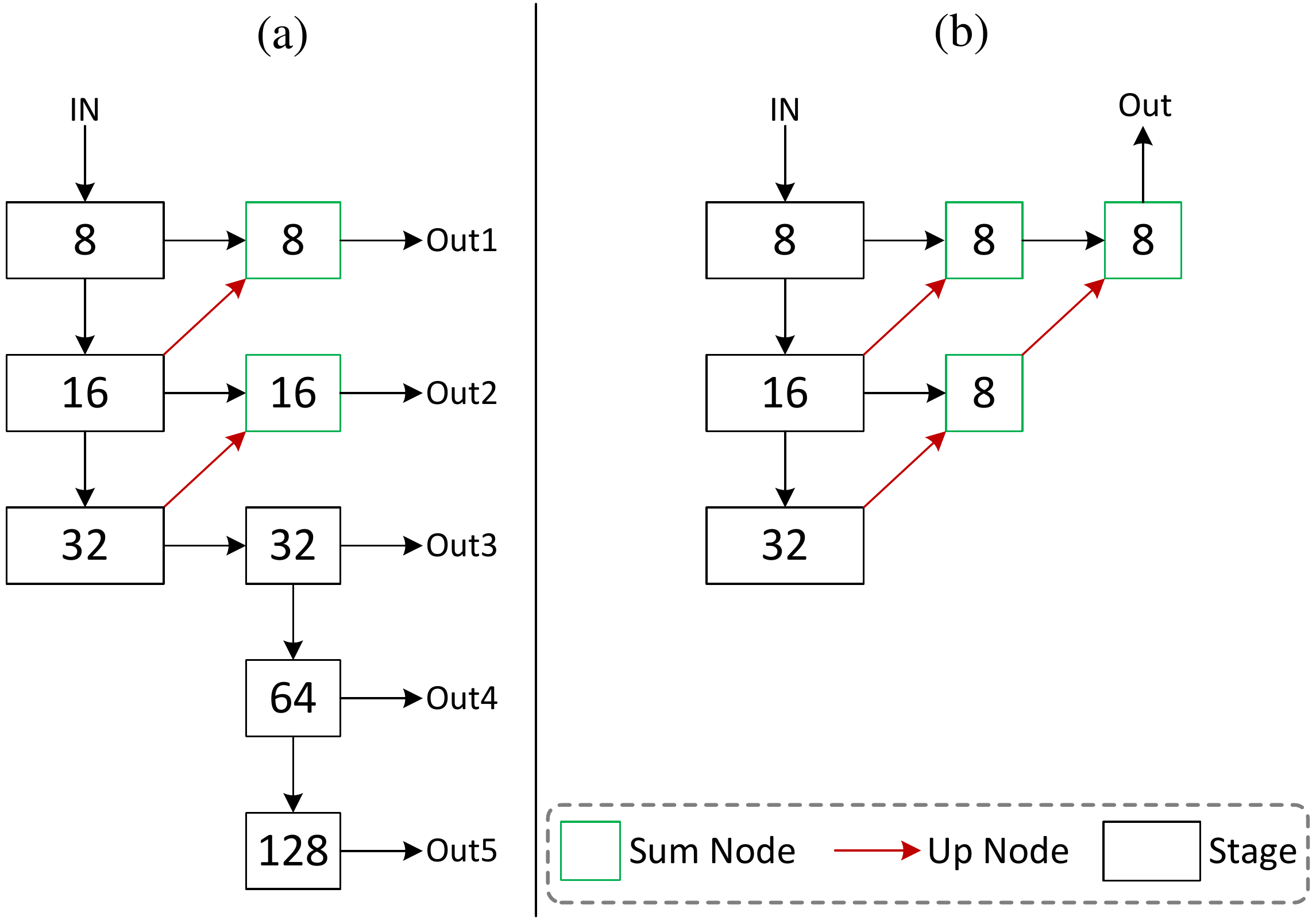}
  \caption{(a) In the first experiment, the network outputs multi-resolution detection layers. (b) In the second experiment, the network is equipped with two up-convolutional networks and the skip connections from lower layers to the output to allow for a single-resolution detection layer. The number in the box denotes the stride.} 
  \label{fig:model_diagrams} 
\end{figure}

\subsection{Multi-resolution Detection Improves Precision}
As shown in Table~\ref{tab:sota}, the proposed MR-CenterNet further improves object detection accuracy. For instance, at the same network depth (Hourglass-52~\emph{vs.}~ResNet-50), MR-CenterNet improves the AP of objects by $4.8\%$. And thanks to the strong generality of the framework of MR-CenterNet, we are able to apply stronger backbones for CenterNet. 

We also design two comparative experiments to confirm the contribution of multi-resolution detection in MR-CenterNet. For the first experiment, it is a control experiment, so we use the default network. For the second experiment, we augment the network with two up-convolutional networks and the skip connections from lower layers to the output to allow for a higher-resolution output. The resolution of the output layer is $1/8$ times of the input image. Therefore, all the objects are detected in single-resolution detection layer. The diagrams of the two network structures are shown in Figure~\ref{fig:model_diagrams}. Table~\ref{model_diagrams} reports the detection results of the two experiment on the MS COCO validation dataset, MR-CenterNet with multi-resolution detection layers achieves higher accuracy. This is because the multi-resolution detection structure provides richer receptive field for detecting objects with different scales, which helps to improve the detection accuracy.

\begin{table}[!tb]
\tiny
\centering
\caption{The detection results using different network structures on the MS COCO validation dataset.}
\resizebox{.48\textwidth}{!}{
\begin{tabular}{l|c|c|c}
\hline
Method & Backbone & Structure & AP\\
\hline
\hline
MR-CenterNet & R-50 & Figure~\ref{fig:model_diagrams} (a)  & 45.7\\
MR-CenterNet & R-50 & Figure~\ref{fig:model_diagrams} (b) & 40.9\\
\hline
\end{tabular}
}
\label{model_diagrams}
\end{table}

\subsection{Real-time CenterNet}

\begin{table}[tb]
\centering
\small
\caption{We report the inference speed of real-time CenterNet \textit{vs.} other typical detectors on the MS-COCO dataset. 'RT': real-time, 'AP$_{val}$': AP on the MS coco validation dataset, 'AP$_{val}$': AP on the MS coco test-dev dataset. For fair comparison, here we follow FCOS~\cite{tian2020fcos} to measure FCOS-RT, YOLOv3 and our real-time CenterNet in an end-to-end manner, which is from prepossessing to the final output boxes. Our CenterNet achieves a good trade-off between accuracy and speed.}
\begin{tabular}{l|c|c|c|c}
    \hline
    Method & Backbone & FPS & AP$_{val}$ & AP$_{test}$  \\
    \hline
    YOLOv3~\cite{redmon2018yolov3} & Darknet-53 & 26 & $-$  & 33.0 \\
    Objects as Points~\cite{zhou2019objects} &  DLA-34  & 52 & 37.4 & 37.3 \\
    FCOS-RT~\cite{tian2020fcos} & R-50 & 38 & 40.2 & 40.2 \\
    CPNDet~\cite{duan2020corner} & DLA-34 & 26.2 & 41.6 & 41.8\\
    \hline
    \hline
    SR-CenterNet & HG-104 & 5.1 & 44.7 & 44.9 \\
    SR-CenterNet & HG-52 & 6.8 & 41.3 & 41.6 \\
    MR-CenterNet & R-50 & 14.5 & 45.7 & 46.4\\
    CenterNet-RT & R-50 & 30.5 & 43.2 & 43.6 \\
    \hline
\end{tabular}
\label{table:real_time}
\end{table}

We also design a real-time version of CenterNet, which is called CenterNet-RT. CenterNet-RT is based on the multi-resolution detection framework of CenterNet (show in Figure~\ref{fig:py_centernet}), we apply the following tricks to speed up CenterNet: (i). Reduce the resolution of the input image, which from $800$ to $512$ and $1333$ to $736$ for shorter side and the maximum longer side, respectively; (ii). Inspired by FCOS-RT~\cite{tian2020fcos} and BlendMask-RT~\cite{chen2020blendmask}, we remove $P_6$ and $P_7$ levels to further save the calculation, since the low resolution input image makes the higher feature levels $P_6$ and $P_7$ less important. (iii). Remove the corner and center keypoint detection heatmap. For CenterNet-RT, we only use the regression results of corner and center keypoint as the final predictions of the keypoints. This saves a lot of time which spent on the cascade corner pooling and finding the closest keypoints on the heatmap. (iv). During training, replace the SGD optimizer with the AdamW~\cite{loshchilov2017decoupled} optimizer and increase the number of training epochs to $32$. We test the inference speed of our method on an NVIDIA Tesla-V100 GPU. We also follow FCOS~\cite{tian2020fcos} to measure FCOS-RT, YOLOv3 and our real-time CenterNet in an end-to-end manner, which is from prepossessing to the final output boxes.

\begin{table}[!tb]
\centering
\caption{Comparison of the average false discovery rates ($\%$) of CornerNet and CenterNet on the MS-COCO validation dataset. The results suggest CenterNet avoids a large number of incorrect bounding boxes, especially for small incorrect bounding boxes.}
\resizebox{.48\textwidth}{!}{
\begin{tabular}{l|c|ccccccc}
\hline
 Method & Backbone & AF & AF$_{5}$ & AF$_{25}$ & AF$_{50}$ & AF$_{\mathrm{S}}$ & AF$_{\mathrm{M}}$ & AF$_{\mathrm{L}}$\\
\hline
\hline
 CornerNet & HG-52 & 40.4 & 35.2 & 39.4 & 46.7 & 62.5 & 36.9 & 28.0 \\
 SR-CenterNet & HG-52 & \textbf{35.1} & \textbf{30.7} & \textbf{34.2} & \textbf{40.8} & \textbf{53.0} & \textbf{31.3} & \textbf{24.4} \\
\hline
\hline
 CornerNet & HG-104 & 37.8 & 32.7 & 36.8 & 43.8 & 60.3 & 33.2 & 25.1 \\
 SR-CenterNet & HG-104 & \textbf{32.4} & \textbf{28.2} & \textbf{31.6} & \textbf{37.5} & \textbf{50.7} & \textbf{27.1} & \textbf{23.0} \\
\hline
\end{tabular}}
\label{table:FDR2}
\end{table}

\begin{table*}[tb]
\small   
\centering
\caption{Ablation study on the major components of CenterNet511-52 on the MS-COCO validation dataset. The CRE denotes central region exploration, the CTP denotes center pooling, and the CCP denotes cascade corner pooling.}
\begin{tabular}{*{3}{p{0.785cm}<{\centering}}|*{6}{p{0.71cm}<{\centering}}|*{6}{p{0.71cm}<{\centering}}}
\hline
CRE & CTP & CCP & AP & AP$_{50}$ & AP$_{75}$ & AP$_\mathrm{S}$ & AP$_\mathrm{M}$ & AP$_\mathrm{L}$ & AR$_1$ & AR$_{10}$ & AR$_{100}$ & AR$_\mathrm{S}$ & AR$_\mathrm{M}$ & AR$_\mathrm{L}$\\
\hline
\hline
 &  &  & 37.6 & 53.3 & 40.0 & 18.5 & 39.6 & 52.2 & 33.7 & 52.2 & 56.7 & 37.2 & 60.0 & 74.0 \\
 \hline
 &  & \checkmark & 38.3 & 54.2 & 40.5 & 18.6 & 40.5 & 52.2 & 34.0 & 53.0 & 57.9 & 36.6 & 60.8 & 75.8 \\
 \hline
\checkmark &  &  & 39.9 & 57.7 & 42.3 & 23.1 & 42.3 & 52.3 & 33.8 & 54.2 & 58.5 & 38.7 & 62.4 & 74.4 \\
\hline
\checkmark & \checkmark &  & 40.8 & 58.6 & 43.6 & 23.6 & 43.6 & 53.6 & 33.9 & 54.5 & 59.0 & 39.0 & 63.2 & 74.7 \\
\hline
\checkmark & \checkmark & \checkmark & \textbf{41.3} & \textbf{59.2} & \textbf{43.9} & \textbf{23.6} & \textbf{43.8} & \textbf{55.8} & \textbf{34.5} & \textbf{55.0} & \textbf{59.2} & \textbf{39.1} & \textbf{63.5} & \textbf{75.1} \\
\hline
\end{tabular}
\label{ablation}
\end{table*}

The results are shown in Table~\ref{table:real_time}. CenterNet-RT achieves a good trade-off between accuracy and speed and is still competitive among other typical methods. In our conference version, CenterNet (\ie, SR-CenterNet) performs slow, which the inference speed is less than $7$ FPS. In this paper, we equip CenterNet with a pyramid structure and detect objects in multi-resolution feature layers, and  the improvement is significant for speed and accuracy. With ResNet-50, MR-CenterNet achieves $45.7\%$ AP at $14.5$ FPS. We further propose the CenterNet-RT on the basis of MR-CenterNet, achieving results of $43.2\%$ AP at $30.5$ FPS. This accuracy is competitive with SR-CenterNet (HG-104) but the inference speed is $\sim 6$ times faster.

\subsection{Incorrect Bounding Box Reduction}
The AP~\cite{lin2014microsoft} metric reflects how many high quality object bounding boxes (usually $\mathrm{IoU} \geqslant 0.5$) a network can predict, but cannot directly reflect how many incorrect object bounding boxes (usually $\mathrm{IoU} \ll 0.5$) a network generates. The AF rate is a suitable metric, which reflects the proportion of the incorrect bounding boxes. Table~\ref{table:FDR2} shows the AF rates for CornerNet and CenterNet. CornerNet generates many incorrect bounding boxes even at $\mathrm{IoU} = 0.05$ threshold, \ie, CornerNet511-52 and CornerNet511-104 obtain $35.2\%$ and $32.7\%$ AF rate, respectively. On the other hand, CornerNet generates more small incorrect bounding boxes than medium and large incorrect bounding boxes, which reports $62.5\%$ for CornerNet511-52 and $60.3\%$ for CornerNet511-104, respectively. Our CenterNet decreases the AF rates at all criteria via exploring central regions. For instance, CenterNet511-52 and CenterNet511-104 decrease $\mathrm{AF_{5}}$ by both $4.5\%$. In addition, the AF rates for small bounding boxes decrease the most, which are $9.5\%$ by CenterNet511-52 and $9.6\%$ by CenterNet511-104, respectively. This is also the reason why the AP improvement for small objects is more prominent.

\subsection{Ablation Study}
Our work has contributed three components, including central region exploration, center pooling and cascade corner pooling. To analyze the contribution of each individual component, an ablation study is given here. The baseline is CornerNet511-52~\cite{law2018cornernet}. We add the three components to the baseline one by one and follow the default parameter setting detailed in Section~\ref{sec:setting}. The results are given in Table~\ref{ablation}. 

\noindent\textbf{Central region exploration.} To understand the importance of the central region exploration (see CRE in the table), we add a center heatmap branch to the baseline and use a triplet of keypoints to detect bounding boxes. For the center keypoint detection, we only use conventional convolutions. As presented in the third row in Table~\ref{ablation}, we improve the AP by $2.3\%$ (from $37.6\%$ to $39.9\%$). However, we find that the improvement for the small objects (that is $4.6$\%) is more significant than that for other object scales. The improvement for large objects is almost negligible (from $52.2\%$ to $52.3\%$). This is not surprising because,  the number of small incorrect bounding boxes are larger and they usually do not contain the center keypoints of objects, which are more likely to benefit from filtering by center keypoints.

\noindent\textbf{Center pooling.} To demonstrate the effectiveness of proposed center pooling, we then add the center pooling module to the network (see CTP in the table). The fourth row in Table~\ref{ablation} shows that center pooling improves the AP by $0.9\%$ (from $39.9\%$ to $40.8\%$). Notably, with the help of center pooling, we improve the AP for large objects by $1.4\%$ (from $52.2\%$ to $53.6\%$), which is much higher than the improvement using conventional convolutions (\ie,~$1.4\%$~\emph{vs.}~$0.1\%$). It demonstrates that our center pooling is effective in detecting center keypoints of objects, especially for large objects. Our explanation is that center pooling can extract richer internal visual patterns, and larger objects contain more accessible internal visual patterns. Figure~\ref{fig:fig7_52} shows the results of detecting center keypoints without/with center pooling. We can see the conventional convolution fails to locate the center keypoint for the cow, but with center pooling, it successfully locates the center keypoint.

\noindent\textbf{Cascade corner pooling.} We replace corner pooling~\cite{law2018cornernet} with cascade corner pooling to detect corners (see CCP in the table). The second row in Table~\ref{ablation} shows the results that we test on the basis of CornerNet511-52. We find that cascade corner pooling improves the AP by $0.7\%$ (from $37.6\%$ to $38.3\%$). The last row shows the results that we test on the basis of CenterNet511-52, which improves the AP by $0.5\%$ (from $40.8\%$ to $41.3\%$). The results of the second row show there is almost no change in the AP for large objects (\ie,~$52.2\%$~\emph{vs.}~$52.2\%$), but the AR is improved by $1.8\%$ (from $74.0\%$ to $75.8\%$). This suggests that cascade corner pooling helps to obtain more internal visual patterns in large objects, but too rich visual patterns may interfere with its perception for the boundary information, leading to many inaccurate bounding boxes. After equipping with our CenterNet, the inaccurate bounding boxes are effectively suppressed, which improves the AP for large objects by $2.2\%$ (from $53.6\%$ to $55.8\%$). Figure~\ref{fig:fig7_62} shows the result of detecting corners with corner pooling or cascade corner pooling. We can see that cascade corner pooling can successfully locate a pair of corners for the cat on the left while corner pooling cannot.

\begin{table}[!tb]
\small
\centering
\caption{Error analysis of center keypoints via using ground-truth. we replace the predicted center keypoints with the ground-truth values, the results suggest there is still room for improvement in detecting center keypoints.}
\resizebox{.48\textwidth}{!}{
\begin{tabular}{l|c|cccccc}
\hline
Method & Backbone & AP & AP$_{50}$ & AP$_{75}$ & AP$_{\mathrm{S}}$ & AP$_{\mathrm{M}}$ & AP$_{\mathrm{L}}$\\
\hline
\hline
SR-CenterNet w/o GT & HG-52 & 41.3 & 59.2 & 43.9 & 23.6 & 43.8 & 55.8 \\
SR-CenterNet w/ GT & HG-52 & \textbf{56.5} & \textbf{78.3} & \textbf{61.4} & \textbf{39.1} & \textbf{60.3} & \textbf{70.3} \\
\hline
\hline
SR-CenterNet w/o GT & HG-104 & 44.8 & 62.4 & 48.2 & 25.9 & 48.9 & 58.8\\
SR-CenterNet w/ GT & HG-104 & \textbf{58.1} & \textbf{78.4} & \textbf{63.9} & \textbf{40.4} & \textbf{63.0} & \textbf{72.1} \\
\hline
\end{tabular}
}
\label{Error}
\end{table}

\subsection{Error Analysis} The exploration of visual patterns within each bounding box depends on the center keypoints. In other words, once a center keypoint is missed, the proposed CenterNet would miss the visual patterns within the bounding box. To understand the importance of center keypoints, we replace the predicted center keypoints with the ground-truth values and evaluate performance on the MS-COCO validation dataset. Table~\ref{Error} shows that using the ground-truth center keypoints improves the AP from $41.3\%$ to $56.5\%$ for CenterNet511-52 and from $44.8\%$ to $58.1\%$ for CenterNet511-104, respectively. APs for small, medium and large objects are improved by $15.5\%$, $16.5\%$, and $14.5\%$ for CenterNet511-52 and $14.5\%$, $14.1\%$, and $13.3\%$ for CenterNet511-104, respectively. 

\section{Conclusions}
\label{sec:conclusions}
In this paper, we propose CenterNet, a new bottom-up object detection approach that detects objects using a triplet, including one center keypoint and two corners. Our approach addresses the problem that the traditional bottom-up approach lack an additional look into the cropped regions by exploring the visual patterns within each proposed region with minimal costs. Besides, we also extend the CenterNet on a framework that has pyramid structure to allow for better detecting multi-scale objects. The experimental results show CenterNet outperforms all existing bottom-up approaches by a large margin and is competitive among the top-down approaches especially in recall ability. We also design some real-time models for CenterNet, which achieves a good trade-off between accuracy and speed.

The most important take-away is that we prove that bottom-up approaches are more flexible in locating objects with arbitrary geometries, and an additional look into each proposed region is necessary for improving precision. We hope that CenterNet could attract more attention to promote a further exploration of bottom-up approaches.
 

%




\ifCLASSOPTIONcaptionsoff
  \newpage
\fi


\bibliographystyle{utils/ieee_fullname}
\bibliography{egbib}

%








\end{document}